\documentclass[twocolumn]{IEEEtran}
   
\usepackage{multirow}
\usepackage{amsmath}
\usepackage{bm}
\usepackage{booktabs} 
\usepackage{wasysym}
\usepackage{graphicx} 
\usepackage{dsfont}
\usepackage{bbm}
\usepackage{array,color}
\usepackage{subfigure}
\usepackage{verbatim}
\usepackage{amssymb}
\usepackage{ifthen}
\usepackage{url}
\usepackage{microtype}

\usepackage{ifthen}

\definecolor{FanColor}{rgb}{0.8,0,0.8}
\newcommand{\fan}[1]{{\color{FanColor}[Fan: #1]}}

\definecolor{XingjiaColor}{rgb}{0.0,0.1,0.9}
\newcommand{\xingjia}[1]{{\color{XingjiaColor}[Xingjia: #1]}}

\newcommand{\warning}[1]{{\it\color{red} #1}}
\newcommand{\toremove}[1]{{\it\color{red} (To remove) #1}}
\newcommand{\note}[1]{{\it\color{blue} #1}}
\newcommand{\nothing}[1]{}

{
\renewcommand{\fan}[1]{}
\renewcommand{\xingjia}[1]{}

\renewcommand{\warning}[1]{}
\renewcommand{\toremove}[1]{}
\renewcommand{\note}[1]{}
\renewcommand{\nothing}[1]{}
}{}

\newcommand{\citep}[1]{\cite{#1}}

\newcommand{\figtext}[1]{{\footnotesize #1}}
\newcommand{\smallfigtext}[1]{{\scriptsize #1}}

 \newcommand{\loss}{\mathcal{L}}
\newcommand{\contentloss}{\loss_{c}}
\newcommand{\styleloss}{\loss_{s}}

\newcommand{\msamodule}{\mathcal{F}_\mathrm{MSA}}
\newcommand{\ffnetwork}{\mathcal{F}_\mathrm{FFN}}
\newcommand{\pemodule}{\mathcal{P}}
\newcommand{\lpemodule}{\mathcal{P_L}}
\newcommand{\capemodule}{\mathcal{P_{CA}}}

\newcommand{\featureembedding}{\mathcal{E}}
\newcommand{\encodingfunction}{\mathcal{F}_\mathrm{pos}}
\newcommand{\averagepooling}{\mathrm{AvgPool}}
\newcommand{\concatenation}{\mathrm{Concat}}
\newcommand{\attention}{\mathrm{Attention}}
\newcommand{\convolution}{\mathrm{Conv}}

\newcommand{\relu}{\mathrm{ReLU}}
\newcommand{\upsampling}{\mathrm{Upsample}}

\newcommand{\inputsequence}{Z}
\newcommand{\inputcontentsequence}{\inputsequence_c}
\newcommand{\inputstylesequence}{\inputsequence_s}
\newcommand{\encodedsequence}{Y}
\newcommand{\encodedcontentsequence}{\encodedsequence_c}
\newcommand{\encodedstylesequence}{\encodedsequence_s}
\newcommand{\outputsequence}{X}

\newcommand{\inputcontent}{I_c}
\newcommand{\inputstyle}{I_s}
\newcommand{\outputimage}{I_o}
\newcommand{\layernumber}{N_l}

\newcommand{\generator}{G}
\newcommand{\distance}{d}

 \usepackage{engord}
 
 \usepackage{floatrow}
\newfloatcommand{capbtabbox}{table}

\ifCLASSOPTIONcompsoc
  \usepackage[nocompress]{cite}
\else
  \usepackage{cite}
\fi

\ifCLASSINFOpdf
\else
\fi

\begin{document}

\title{StyTr$^2$:Image Style Transfer with Transformers}

\author{Yingying~Deng, Fan~Tang, Weiming~Dong~\IEEEmembership{Member,~IEEE}, Chongyang~Ma, Xingjia~Pan, Lei Wang, Changsheng~Xu, ~\IEEEmembership{Fellow,~IEEE}
\IEEEcompsocitemizethanks{
\IEEEcompsocthanksitem Yingying Deng, Weiming Dong and Changsheng Xu are with School of Artificial Intelligence, University of Chinese Academy of Sciences, Beijing, China and NLPR, Institute of Automation, Chinese Academy of Sciences, Beijing, China (e-mail: \{dengyingying2017, weiming.dong, changsheng.xu\}@ia.ac.cn).
\IEEEcompsocthanksitem Fan Tang is with Jilin University, Changchun, China (e-mail: tangfan@jlu.edu.cn)
\IEEEcompsocthanksitem Xingjia Pan is with YouTu Lab, Tencent, Shanghai, China (e-mail: xjia.pan@gmail.com)
\IEEEcompsocthanksitem Chongyang Ma is with Kuaishou Technology, Beijing, China (e-mail: chongyangma@kuaishou.com).
}
}


\IEEEtitleabstractindextext{%
\begin{abstract}
The goal of image style transfer is to render an image with artistic features guided by a style reference while maintaining the original content. 
Owing to the locality in convolutional neural networks (CNNs), extracting and maintaining the global information of input images is difﬁcult. 
Therefore, traditional neural style transfer methods face biased content representation. 
To address this critical issue, we take long-range dependencies of input images into account for image style transfer by proposing a transformer-based approach called StyTr$^2$. 
In contrast with visual transformers for other vision tasks, StyTr$^2$ contains two different transformer encoders to generate domain-specific sequences for content and style, respectively. 
Following the encoders, a multi-layer transformer decoder is adopted to stylize the content sequence according to the style sequence.
We also analyze the deficiency of existing positional encoding methods and propose the content-aware positional encoding (CAPE), which is scale-invariant and more suitable for image style transfer tasks. Qualitative and quantitative experiments demonstrate the effectiveness of the proposed StyTr$^2$ compared with state-of-the-art CNN-based and flow-based approaches.
Code and models are available at \url{https://github.com/diyiiyiii/StyTR-2}. 
\end{abstract}

 }

\maketitle

\IEEEdisplaynontitleabstractindextext

\IEEEpeerreviewmaketitle

\section{Introduction}
\label{sec:Introduction}
Image style transfer is an interesting and practical research topic that can render a content image using a referenced style image.
Based on texture synthesis, traditional style transfer methods~\cite{efros:2001:image,bruckner:2007:style} can generate vivid stylized images, but are computationally complex due to the formulation of stroke appearance and painting process.
Afterward, researchers focus on neural style transfer based on convolutional neural networks (CNNs).
Optimization-based style transfer methods~\cite{gatys:2016:image, li:2017:demystifying, risser:2017:stable} render the input content images with learned style representation iteratively.
Following the encoder-transfer-decoder pipeline, arbitrary style transfer networks~\cite{Huang:2017:Arbitrary,li:2017:universal,wang:2020:diversified,li:2018:closed,2021:Lin:Drafting,2020:An:Real,2019:Lu:A,2019:An:Ultrafast,2020:Wang:Collaborative} are optimized by aligning second-order statistics of content images to style images and can generate stylized results in a feed-forward manner efficiently.
However, these methods cannot achieve satisfactory results in some cases due to the limited ability to model the relationship between content and style.
To overcome this issue, several recent methods \cite{park:2019:arbitrary,deng:2020:arbitrary,deng:2021:arbitrary,yao:2019:attention,liu:2021:adaattn} apply a self-attention mechanism for improved stylization results.

\newcommand\teaserfigurewidth{0.23}
\newcommand\teasertextvspace{-2pt}

\begin{figure}
\setlength{\abovecaptionskip}{2mm}
\centering
\begin{minipage}{\teaserfigurewidth\linewidth}
\centering
\begin{minipage}{\linewidth}
    \includegraphics[width=\linewidth]{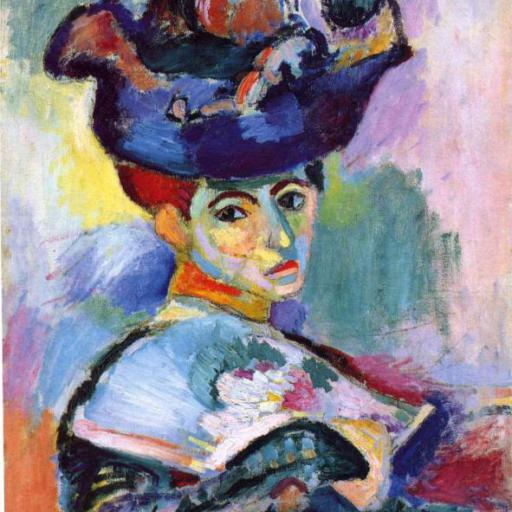}
\end{minipage}
\vspace{\teasertextvspace}
\centering
\smallfigtext{\emph{Woman with a Hat}, Henri Matisse, 1905.}
\end{minipage}
\begin{minipage}{\teaserfigurewidth\linewidth}
\centering
\begin{minipage}{\linewidth}
    \includegraphics[width=\linewidth]{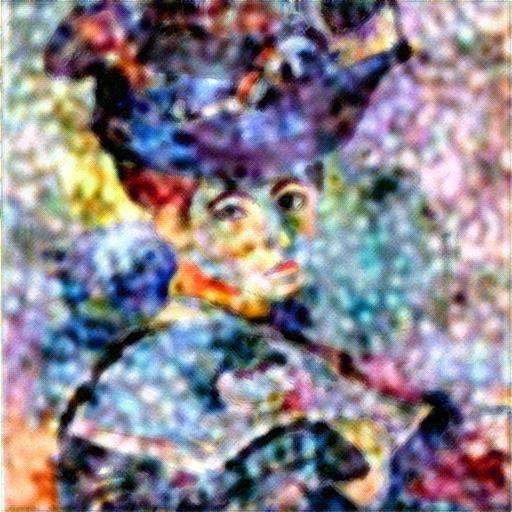}
\end{minipage}
\vspace{\teasertextvspace}
\begin{minipage}{\linewidth}
\centering
    \smallfigtext{conv1\_1}
\end{minipage}

\vspace{7pt}
\begin{minipage}{\linewidth}
    \includegraphics[width=\linewidth]{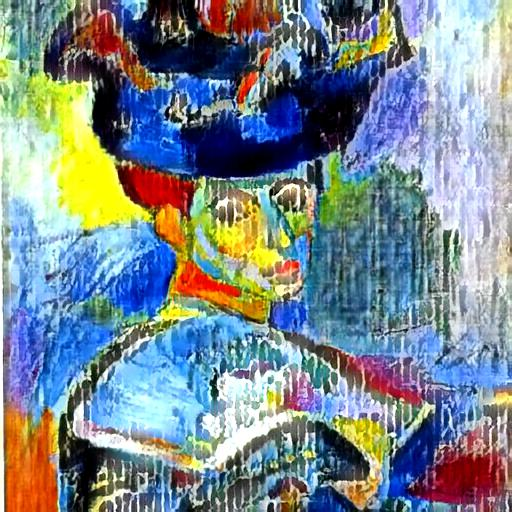}
\end{minipage}
\vspace{\teasertextvspace}
\begin{minipage}{\linewidth}
\centering
    \smallfigtext{layer\_1}
\end{minipage}
\end{minipage}
\begin{minipage}{\teaserfigurewidth\linewidth}
\centering
\begin{minipage}{\linewidth}
    \includegraphics[width=\linewidth]{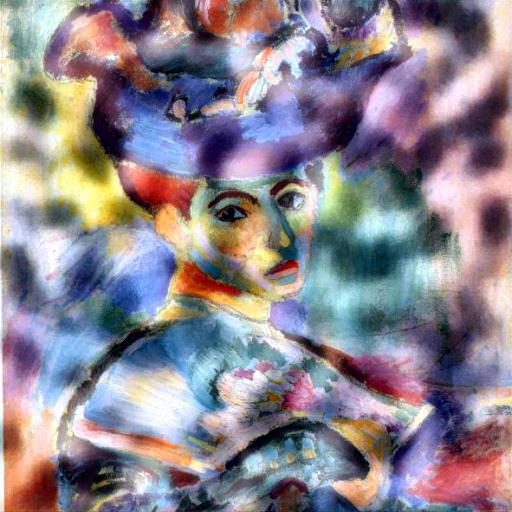}
\end{minipage}
\vspace{\teasertextvspace}
\begin{minipage}{\linewidth}
\centering
    \smallfigtext{\{conv1\_1,conv2\_1\}}
\end{minipage}

\vspace{7pt}
\begin{minipage}{\linewidth}
    \includegraphics[width=\linewidth]{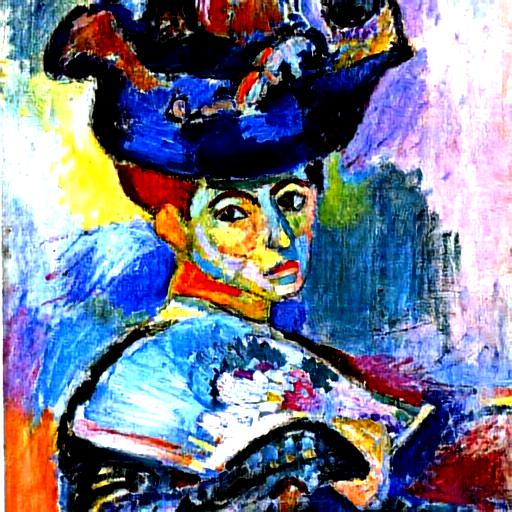}
\end{minipage}
\vspace{\teasertextvspace}
\begin{minipage}{\linewidth}
\centering
    \smallfigtext{layer\_2}
\end{minipage}
\end{minipage}
\begin{minipage}{\teaserfigurewidth\linewidth}
\centering
\begin{minipage}{\linewidth}
    \includegraphics[width=\linewidth]{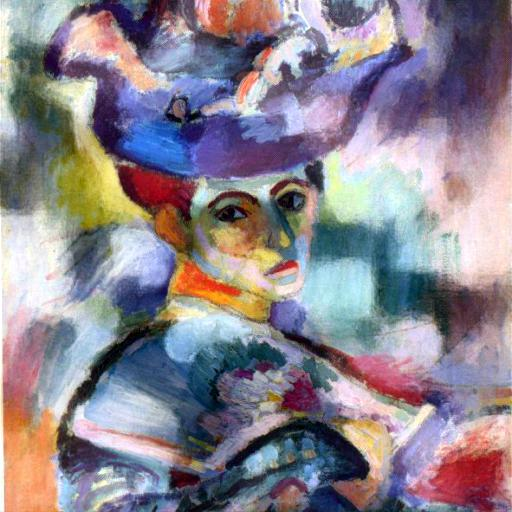}
\end{minipage}
\vspace{\teasertextvspace}
\begin{minipage}{\linewidth}
\centering
    \smallfigtext{\{conv1\_1,...,conv4\_1\}}
\end{minipage}

\vspace{7pt}
\begin{minipage}{\linewidth}
    \includegraphics[width=\linewidth]{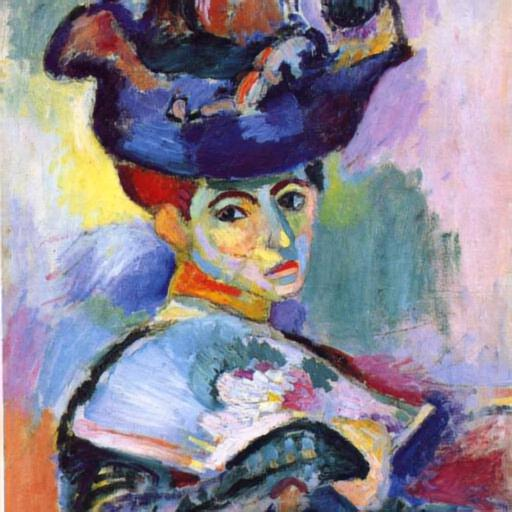}
\end{minipage}
\vspace{\teasertextvspace}
\begin{minipage}{\linewidth}
\centering
    \smallfigtext{layer\_3}
\end{minipage}
\end{minipage}
\begin{minipage}{0.03\linewidth}
    \centering
    \figtext{(a)}

    \vspace{55pt}

    \figtext{(b)}
\end{minipage}
\caption{Comparisons of intermediate layers using the leftmost image as the input content and the style reference in a style transfer task. (a) Feature visualizations of a pretrained VGG based on \cite{gatys:2016:image}.(b) Feature visualizations of our transformer decoder.}
\label{fig:cnnvis}
\end{figure}
 
The aforementioned style transfer methods utilize CNNs to learn style and content representations.
Owing to the limited receptive field of convolution operation, CNNs cannot capture long-range dependencies without sufficient layers. 
However, the increment of network depth could cause the loss of feature resolution and fine details \cite{jiang:2021:transgan}. 
The missing details can damage the stylization results in aspects of content structure preservation and style display.
As shown in Figure \ref{fig:cnnvis}(a), some details are omitted in the process of convolutional feature extraction.
An et al \cite{an:2021:artflow} recently show that typical CNN-based style transfer methods are biased toward content representation by visualizing the content leak of the stylization process, i.e., after repeating several rounds of stylization operations, the extracted structures of input content will change drastically.

With the success of transformer~\cite{vaswani:2017:attention} in natural language processing (NLP), transformer-based architectures have been adopted in various vision tasks.
The charm of applying transformer to computer vision lies in two aspects.
First, it is free to learn the global information of the input with the help of the self-attention mechanism. Thus, a holistic understanding can be easily obtained within each layer.
Second, the transformer architecture models relationships in input shapes~\cite{paul:2021:vision}, and different layers extract similar structural information~\cite{raghu:2021:do} (see Figure \ref{fig:cnnvis}(b)).
Therefore, transformer has a strong representation capability to capture precise content representation and avoid fine detail missing. Thus, the generated structures can be well-preserved. 

In this work, we aim to eliminate the biased representation issue of CNN-based style transfer methods and propose a novel image \textbf{Sty}le \textbf{Tr}ansfer \textbf{Tr}ansformer framework called \textbf{StyTr$^2$}.
Different from the original transformer, we design two transformer-based encoders in our StyTr$^2$ framework to obtain domain-specific information.
Following the encoders, the transformer decoder is used to progressively generate the output sequences of image patches.
Furthermore, towards the positional encoding methods that are proposed for NLP, we raise two considerations:
(1) different from sentences ordered by logic, the image sequence tokens are associated with semantic information of the image content;
(2) for the style transfer task, we aim to generate stylized images of any resolution.
The exponential increase in image resolution will lead to a significant change of positional encoding, leading to large position deviation and inferior output quality.
In general, a desired positional encoding for vision tasks should be conditioned on input content while being invariant to image scale transformation.
Therefore, we propose a content-aware positional encoding scheme (\textbf{CAPE}) which learns the positional encoding based on image semantic features and dynamically expands the position to accommodate different image sizes.

In summary, our main contributions include: 
\begin{itemize}
\item A transformer-based style transfer framework called StyTr$^2$, to generate stylization results with well-preserved structures and details of the input content image.
\item A content-aware positional encoding scheme that is scale-invariant and suitable for style transfer tasks.
\item Comprehensive experiments showing that StyTr$^2$ outperforms baseline methods and achieves outstanding results with desirable content structures and style patterns.
\end{itemize}


\section{Related Work}
\label{sec:related_work}

\paragraph{Image style transfer}
Gatys et al~\cite{gatys:2016:image} find that hierarchical layers in CNNs can be used to extract image content structures and style texture information and propose an optimization-based method to generate stylized images iteratively.
Some approaches~\cite{johnson:2016:perceptual,li:2016:precomputed} adopt an end-to-end model to achieve real-time style transfer for one specific style.
For more efficient applications, \cite{chen:2017:stylebank,dumoulin:2016:learned,Lin:2021:DAM} combine multiple styles in one model and achieve outstanding stylization results.
More generally, arbitrary style transfer gains more attention in recent years.
Huang et al~\cite{Huang:2017:Arbitrary} propose an adaptive instance normalization (AdaIN) to replace the mean and variance of content with that of style. 
AdaIN is widely adopted in image generation tasks~\cite{karras:2019:stylegan,huang:2018:multimodal,2020:Wang:Collaborative,2020:An:Real,2021:Lin:Drafting} to fuse the content and style features.
Li et al~\cite{li:2017:universal} design a whiten and colorization transformation (WCT) to align the second-order statistics of content and style features.
Moreover, many methods~\cite{wu:2020:Exchangeable,an:2021:artflow,svoboda:2020:two} also aim at promoting the generation effect in the premise of efficiency.
Based on the CNNs model, \cite{deng:2020:arbitrary,deng:2021:arbitrary,park:2019:arbitrary,liu:2021:adaattn,wu:2021:styleformer} introduce self-attention to the encoder-transfer-decoder framework for better feature fusion.
Chen et al ~\cite{chen:2021:iest} propose an Internal-External Style Transfer algorithm (IEST) containing two types of contrastive loss, which can produce a harmonious and satisfactory stylization effect.
However, existing encoder-transfer-decoder style transfer methods cannot handle the long-range dependencies and may lead to missing details.

\begin{figure*}
\setlength{\abovecaptionskip}{2mm}
\centering
\includegraphics[width=0.95\linewidth]{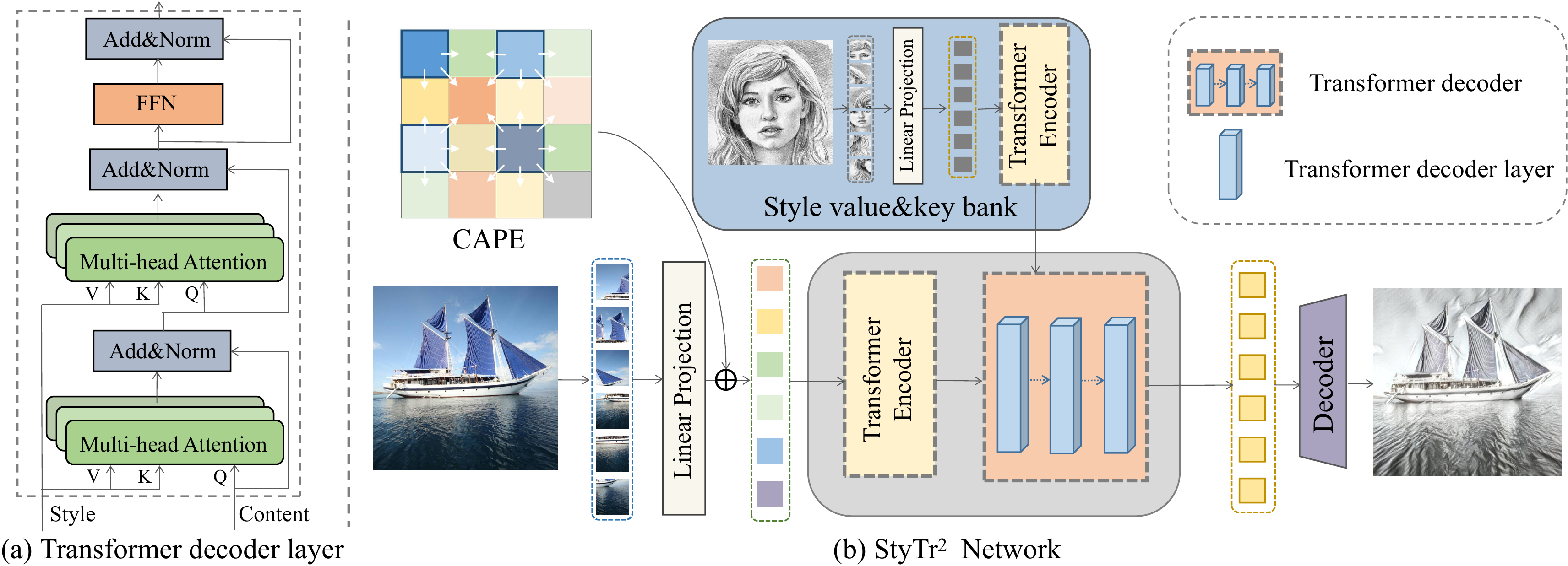}
\caption{Overall pipeline of our StyTr$^2$ framework. We split the content and style images into patches and use a linear projection to obtain patch sequences. Then, the content sequences added with CAPE are fed into a content transformer encoder, while the style sequences are fed into a style transformer encoder. Following the two transformer encoders, a multi-layer transformer decoder is adopted to stylize the content sequences according to the style sequences. Finally, we use a progressive upsampling decoder to obtain the final output.
}
\label{fig:network}
\end{figure*}

\paragraph{Transformer for vision tasks}
As an alternative to recurrent and convolutional neural networks, transformer~\cite{vaswani:2017:attention} is first proposed for machine translation tasks and has been widely used in various NLP tasks~\cite{radford:2019:language,brown:2020:language,devlin:2018:bert,liu:2019:roberta,radford:2018:improving, dai:2019:style}.
Inspired by the breakthrough of transformer in NLP, many researchers have developed vision transformers for various image/video related tasks~\cite{xu:2022:transformers}, including object detection~\cite{carion:2020:dert,zhu:2020:deformabledetr, dai:2020:up}, semantic segmentation~\cite{zheng:2021:setr,wang:2020:end}, image classification~\cite{dosovitskiy2021vit,wu:2020:visual,chen:2020:generative, liu:2021:SwinTransformer,xuL:2021:evo}, image processing and generation~\cite{chen:2020:pre,chen:2020:generative,jiang:2021:transgan}.
Compared with fully convolutional networks, transformer-based networks can capture long-term dependencies of the input image by using self-attention mechanisms.
In this paper, we introduce transformer-based structures for style transfer tasks which can be seen as sequence-to-sequence generation of image patches.

\paragraph{Positional encoding}
Positional encoding is commonly used in transformer-based models to provide position information.
There are two types of positional encoding are used: \emph{functional} and \emph{parametric} positional encoding.
Functional positional encoding is calculated by pre-defined functions, such as sinusoidal functions~\cite{vaswani:2017:attention}.
Parametric positional encoding is learned via model training~\cite{devlin:2018:bert}.
To ensure translational-invariance for the transformers, relative positional encoding~\cite{shaw:2018:self,yang:2019:xlnet,raffel:2019:exploring,he:2020:deberta} considers the distance between tokens in the image sequence.
\cite{xu:2020:positional} and \cite{2020:Islam:How} further include positional encoding in CNN-based models as spatial inductive.
In this paper, we propose a content-aware positional encoding mechanism that is scale-invariant and more suitable for image generation tasks.

\section{Our Method}
\label{sec:Method}

To leverage the capability of transformers to capture long-range dependencies of image features for style transfer, we formulate the problem as a sequential patch generation task.
Given a content image $\inputcontent \in \mathbb{R}^{ H \times W \times 3}$ and a style image $I_s \in \mathbb{R}^{ H \times W \times 3}$, we split both images into patches (similar to tokens in NLP tasks) and use a linear projection layer to project input patches into a sequential feature embedding  $\featureembedding$ in a shape of $ L \times C$, where $L = \frac{H \times W}{m \times m}$ is the length of $\featureembedding$, $m=8$ is the patch size and $C$ is the dimension of $\featureembedding$.
The overall structure of our framework is shown in Figure \ref{fig:network}.

\subsection{Content-Aware Positional Encoding}
\label{sec:method_cape}

\begin{figure}
\setlength{\abovecaptionskip}{2mm}
\centering
\includegraphics[width= \linewidth]{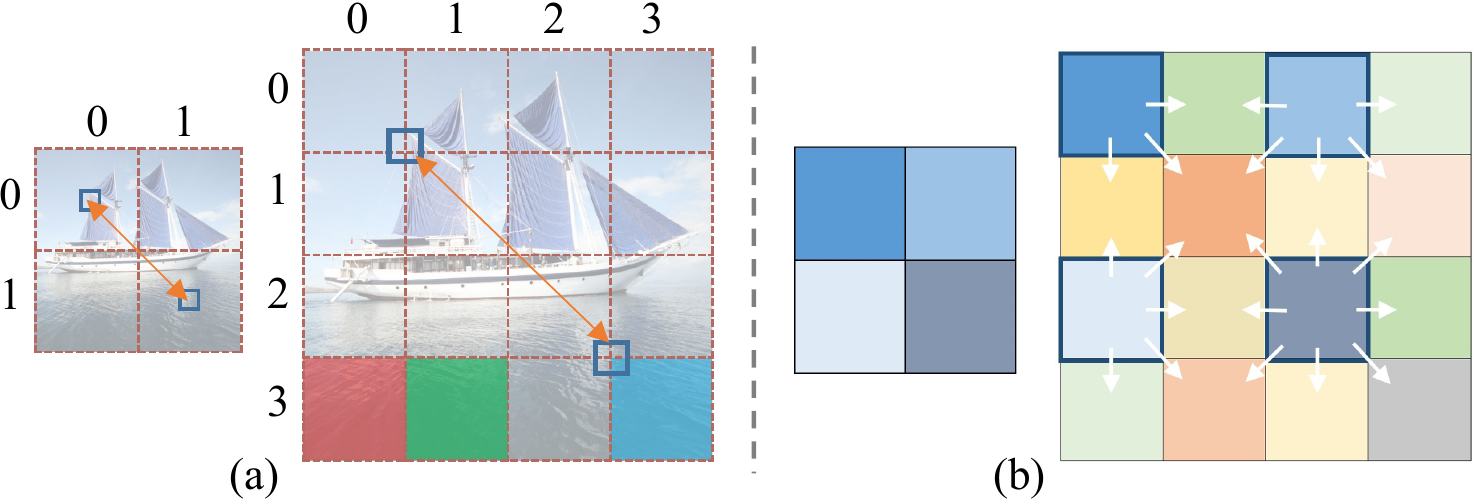}
\caption{Schematic diagram of content-aware positional encoding.
}
\label{fig:position}
\end{figure}

When using a transformer-based model, the positional encoding (\textbf{PE}) should be included in the input sequence to acquire structural information.
According to \cite{vaswani:2017:attention}, the attention score of the $i$-th patch and the $j$-th patch is computed as:
\begin{equation}
\begin{split}
A_{i,j} =& ((\featureembedding_i + \pemodule_i)W_q)^T((\featureembedding_j + \pemodule_j)W_k) \\
=& W_q^T \featureembedding_i^T\featureembedding_j W_k + W_q^T\featureembedding_i^T \pemodule_j W_k \\
&+ W_q^T \pemodule_i^T \featureembedding_j W_k+ W_q^T \pemodule_i^T \pemodule_j W_k,
 \end{split}
\label{fun:att}
\end{equation}
where $W_q$ and $W_k$ are parameter matrices for query and key calculation, and $\pemodule_i$ presents the $i$-th one-dimensional PE.
In 2D cases, the positional relative relation between the patch at a pixel $(x_i, y_i)$ and the patch at a pixel $(x_j, y_j)$ is:
\begin{equation}
\begin{split}
&\pemodule(x_i,y_i)^T \pemodule(x_j,y_j) \\
&= \sum_{k=0}^{\frac{d}{4} -1} [\cos(w_k (x_j - x_i)) + \cos(w_k (y_j - y_i))],
\end{split}
\label{fun:relative}
\end{equation}
where $w_k =1/10000^{2k/128}$, $d = 512$.
The positional relative relation between two patches only depends on their spatial distance.
Accordingly, we raise two important questions.
\emph{First, for an image generation task, should we take image semantics into account when calculating PE?}
Traditional PE is designed for sentences ordered by logic, but image patches are organized based on the content.
We denote the distance between two patches as $\distance( \cdot, \cdot )$.
On the right-hand side of Figure \ref{fig:position}(a), the difference between $\distance((x_0,y_3),(x_1,y_3))$ (the red and green patches) and $\distance((x_0,y_3),(x_3,y_3))$ (the red and cyan patches) should be small because we expect similar content patches to have similar stylization results.
\emph{Second, is the traditional sinusoidal positional encoding still suitable for vision tasks when the input image size expands exponentially?}
As shown in Figure \ref{fig:position}(a), when an image is resized, the relative distance between patches (depicted by small blue rectangles) in the same locations can change dramatically, which may be not suitable for multi-scale methods in vision tasks.

To this end, we propose content-aware positional encoding (CAPE), which is scale-invariant and more suitable for style transfer tasks.
Different from sinusoidal PE which only considers the relative distance of patches, CAPE is conditioned on the semantics of image content.
We assume that using $n \times n$ positional encodings is adequate to represent the semantics of an image.
For an image $I \in \mathbb{R}^{ H \times W \times 3}$, we rescale the fixed $n \times n$ positional encoding to $\frac{H}{m} \times \frac{W}{m}$, as shown in Figure \ref{fig:position}(b).
In this way, various image scales will not influence the spatial relation between two patches.
The CAPE of patch $(x,y)$, namely , $\capemodule{(x,y)}$, is formulated as
\begin{equation}
\begin{split}
\lpemodule &= \encodingfunction(\averagepooling_{n \times n}(\featureembedding)),  \\
\capemodule{(x,y)} &= \sum_{k=0}^{s}\sum_{l=0}^{s}(a_{kl}\lpemodule{(x_k,y_l)}),
\end{split}
\end{equation}
where $\averagepooling_{n \times n}$ is the average pooling function, $\encodingfunction$ is $1 \times 1$ convolution operation used as a learnable positional encoding function, $\lpemodule$ is learnable PE following the sequence $\featureembedding$, $n$ is set to $18$ in our experiments, $a_{kl}$ is the interpolation weight, and $s$ is the number of neighboring patches.
Lastly, we add $\capemodule_i$ to $\featureembedding_i$ as the final feature embedding of the $i$-th patch at a pixel location $(x, y)$.


\subsection{Style Transfer Transformer}
\paragraph{Transformer encoder.}
We capture long-range dependencies of image patches by using transformer based structure to learn sequential visual representations.
Different from other vision tasks~\cite{carion:2020:dert,dai:2020:up, chen:2020:pre}, the input of tjr style transfer task comes from two different domains, corresponding to natural images and artistic paintings, respectively.
Therefore, StyTr$^2$ has two transformer encoders to encode domain-specific features, which are used to translate a sequence from one domain to another in the next stage.

Given the embedding of an input content sequence $ \inputcontentsequence = \{ \featureembedding_{c1} + \capemodule_1, \featureembedding_{c2} + \capemodule_2, ..., \featureembedding_{cL} + \capemodule_L  \}$, we first feed it into the transformer encoder.
Each layer of the encoder consists of a multi-head self-attention module (MSA) and a feed-forward network (FFN).
The input sequence is encoded into query ($Q$), key ($K$), and value ($V$):
\begin{equation}
Q = \inputcontentsequence  W_q , \ \  K =  \inputcontentsequence W_k, \  \  V = \inputcontentsequence W_v, \\
\end{equation}
where $W_q, W_k, W_v \in \mathbb{R}^{C \times d_{head}} $.
The multi-head attention is then calculated by
\begin{equation}
\begin{split}
 \msamodule (Q, K, V) = \concatenation(\attention_1(Q, K, V), \\
 \dots, \attention_N(Q, K, V))W_o,
\end{split}
\end{equation}
where $W_o \in \mathbb{R}^{C \times C}$ are learnable parameters, $N$ is the number of attention heads, and $d_{head} = \frac{C}{N}$.
The residual connections are applied to obtain the encoded content sequence $\encodedcontentsequence$:
\begin{equation}
\begin{split}
\encodedcontentsequence' & = \msamodule(Q, K, V) + Q , \\
\encodedcontentsequence & = \ffnetwork(\encodedcontentsequence') + \encodedcontentsequence', \\
\end{split}
\end{equation}
where $\ffnetwork(Y'_c) = \max(0,Y'_c W_1 + b_1) W_2 + b_2$.
Layer normalization (LN) is applied after each block~\cite{vaswani:2017:attention}.

Similarly, the embedding of an input style sequence $ \inputstylesequence = \{ \featureembedding_{s1} , \featureembedding_{s2} , ..., \featureembedding_{sL} \}$ is encoded into a sequence $\encodedstylesequence$ following the same calculation process, except that positional encoding is not considered because we do not need to maintain structures of the input style in the final output.

\paragraph{Transformer decoder.}
Our transformer decoder is used to translate the encoded content sequence $\encodedcontentsequence$ according to the encoded style sequence $\encodedstylesequence$ in a regressive fashion.
Different from the auto-regressive process in NLP tasks, we take all the sequential patches as input at one time to predict the output.
As shown in Figure \ref{fig:position}(a), each transformer decoder layer contains two MSA layers and one FFN.
The input of our transformer decoder includes the encoded content sequence, i.e., $\hat{Y}_c = \{ Y_{c1} + \capemodule_1, Y_{c2} + \capemodule_2, ..., Y_{cL} + \capemodule_l  \} $, and the style sequence $\encodedstylesequence = \{ Y_{s1} , Y_{s2} , ..., Y_{sL} \}$.
We use the content sequence to generate the query $Q$, and use the style sequence to generate the key $K$ and the value $V$:
\begin{equation}
Q = \hat{Y_c} W_q , \ \  K =  \encodedstylesequence W_k, \  \  V = \encodedstylesequence  W_v. \\
\end{equation}
Then, the output sequence $\outputsequence$ of the transformer decoder can be calculated by
\begin{equation}
\begin{split}
\outputsequence'' &= \msamodule(Q, K, V) + Q, \\
\outputsequence' &= \msamodule(\outputsequence''+ \capemodule ,K,V) + \outputsequence'' ,\\
\outputsequence &= \ffnetwork(\outputsequence') + \outputsequence'.
\end{split}
\end{equation}
Layer normalization (LN) is also applied at the end of each block~\cite{vaswani:2017:attention}.

\paragraph{CNN decoder.}
The output sequence $\outputsequence$ of the transformer is in a shape of $\frac{HW}{64} \times C $.
Instead of directly upsampling the output sequence to construct the final results, we use a three-layer CNN decoder to refine the outputs of the transformer decoder following \cite{zheng:2021:setr}.
For each layer, we expand the scale by adopting a series of operations including $3 \times 3 \  \convolution + \relu + 2 \times \upsampling $.
Finally, we can obtain the final results in a resolution of $H \times W \times 3 $.

\begin{table*}
\centering
\setlength{\tabcolsep}{4.5pt}
\begin{tabular}{c|ccccccccccc}
\toprule
Resolution & Ours & StyleFormer&IEST&AdaAttN&ArtFlow & MCC &MAST&  AAMS & SANet&Avatar&AdaIN \\
\midrule
256 $\times$ 256&0.116 & 0.013&0.065& 0.104 & 0.142 & 0.013 &0.030& 2.074 & 0.015&0.260&0.007 \\
\midrule
512 $\times$ 512 & 0.661& 0.026& 0.092& 0.213 & 0.418 & 0.015  & 0.096&2.173 &0.019& 0.470&0.008 \\
\bottomrule
\end{tabular}
\vspace{-5pt}
\caption{Average inference time (in seconds) of different methods at two output resolutions.}
\label{tab:time}
\vspace{-7pt}
\end{table*}

\subsection{Network Optimization}
The generated results should maintain the original content structures and the reference style patterns.
Therefore, we construct two different perceptual loss terms to measure the content difference between the output image $\outputimage$ and the input content image $\inputcontent$, as well as the style difference between $\outputimage$ and the input style reference $\inputstyle$.

We use feature maps extracted by a pretrained VGG model to construct the content loss and the style loss following~\cite{Huang:2017:Arbitrary,an:2021:artflow}.
The content perceptual loss $\contentloss$ is defined as
\begin{equation}
\begin{split}
\contentloss = \frac{1}{\layernumber}\sum_{ i=0 }^{\layernumber} \lVert \phi_i(\outputimage) - \phi_i(\inputcontent)  \lVert_2,
\end{split}
\label{fun:contentloss}
\end{equation}
where $\phi_i(\cdot)$ denotes features extracted from the $i$-th layer in a pretrained VGG19 and $\layernumber$ is the number of layers.

The style perceptual loss $\styleloss$ is defined as
\begin{equation}
\begin{split}
\styleloss =& \frac{1}{\layernumber}\sum_{ i=0 }^{\layernumber} \lVert \mu( \phi_i(\outputimage)) - \mu (\phi_i(I_{s})) \lVert_2 \\
&+ \lVert \sigma( \phi_i(\outputimage)) - \sigma (\phi_i(I_{s})) \lVert_2,
\end{split}
\label{fun:styleloss}
\end{equation}
where $\mu(\cdot)$ and $\sigma (\cdot)$ denote the mean and variance of extracted features, respectively.

We also adopt identity loss~\cite{park:2019:arbitrary} to learn richer and more accurate content and style representations.
Specifically, we take two of the same content (style) images into StyTr$^2$, and the generated output $I_{cc}(I_{ss})$ should be identical to the input $\inputcontent(\inputstyle)$.
Therefore, we compute two identity loss terms to measure the differences between $\inputcontent(\inputstyle)$ and $I_{cc}(I_{ss})$:
\begin{equation}
\begin{split}
\mathcal{L}_{id1} &= \lVert I_{cc} - \inputcontent  \lVert_2 + \lVert I_{ss} - \inputstyle  \lVert_2, \\
\mathcal{L}_{id2} &= \frac{1}{\layernumber}\sum_{ i=0 }^{\layernumber} \lVert \phi_i(I_{cc}) - \phi_i(\inputcontent)  \lVert_2 +  \lVert \phi_i(I_{ss}) - \phi_i(\inputstyle)  \lVert_2.
\end{split}
\end{equation}
The entire network is optimized by minimizing the following function:
\begin{equation}
\begin{split}
\mathcal{L} = \lambda_{c}\contentloss  + \lambda_{s} \styleloss +\lambda_{id1}\mathcal{L}_{id1}+ \lambda_{id2}\mathcal{L}_{id2}.
\end{split}
\end{equation}
We set $\lambda_{c}$, $\lambda_{s}$, $\lambda_{id1}$, and $\lambda_{id2}$ to $10$, $7$, $50$, and $1$ to alleviate the impact of magnitude differences.

\begin{figure*}
\setlength{\abovecaptionskip}{2mm}
\centering
\includegraphics[width= 0.95\linewidth]{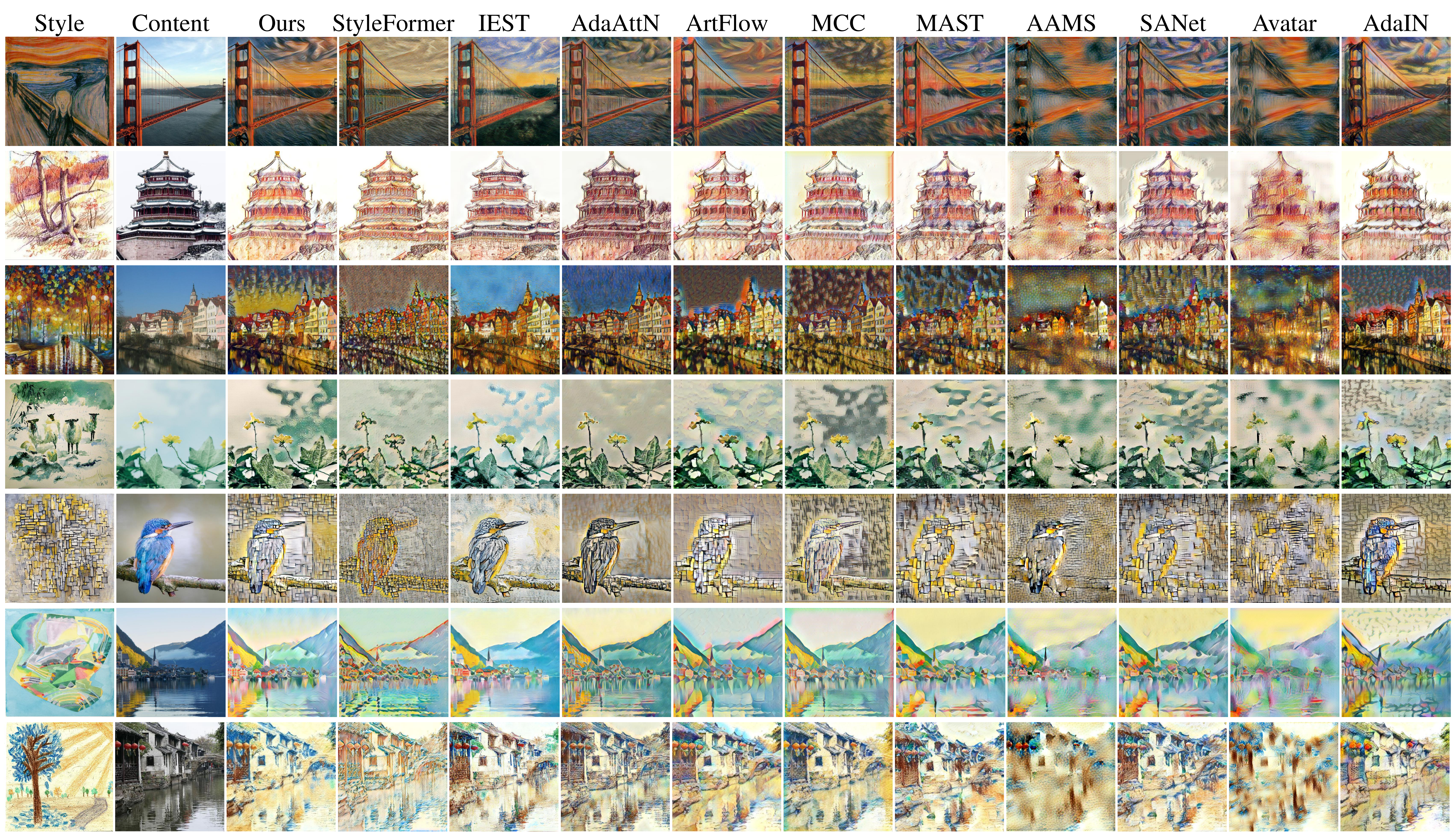}
\caption{Qualitative comparisons of style transfer results using different methods.}
\label{fig:compare}
\end{figure*}


\begin{table*}
\centering
\setlength{\tabcolsep}{4.5pt}
\begin{tabular}{c|ccccccccccc}
\toprule
 &Ours & StyleFormer&IEST&AdaAttN&ArtFlow & MCC &MAST&  AAMS & SANet&Avatar&AdaIN \\
\midrule
$\contentloss$ $\downarrow$ & \textbf{1.91} & 2.86&\underline{1.97}&2.29&2.13 & 2.38 & 2.46&2.44&2.44& 2.84& 2.34 \\
\midrule
$\styleloss$ $\downarrow$ & \underline{1.47} &2.91 &3.47& 2.45& 3.08 & 1.56 &1.55& 3.18 &\textbf{1.18}&2.86& 1.91 \\
\bottomrule
\end{tabular}
\vspace{-5pt}
\caption{Quantitative comparisons. We compute the average content and style loss values of results by different methods to measure how well the input content and style are preserved.The best results are in \textbf{bold} while the second-best results are marked with an \underline{underline}.}
\label{tab:quancomp}
\vspace{-5pt}
\end{table*}

\section{Experiments}
\label{sec:Experiments}

\subsection{Implementation Details}
MS-COCO~\cite{lin:2014:coco} is used as the content dataset and WikiArt~\cite{phillips:2011:wiki} is used as the style dataset.
In the training stage, all the images are randomly cropped into a fixed resolution of $256 \times 256$, while any image resolution is supported at the test time.
We adopt the Adam optimizer~\cite{kingma:2014:adam} and the learning rate is set to $0.0005$ using the warm-up adjustment strategy~\cite{xiong:2020:layer}.
We set the batch size to be $8$ and train our network with $160,000$ iterations.


\subsection{Comparisons with SOTA Methods}

We compare our method with AdaIN~\cite{Huang:2017:Arbitrary}, Avater~\cite{sheng2018avatar}, SANet~\cite{park:2019:arbitrary}, AAMS~\cite{yao:2019:attention}, MAST~\cite{deng:2020:arbitrary}, MCC~\cite{deng:2021:arbitrary}, ArtFlow~\cite{an:2021:artflow}, AdaAttN~\cite{liu:2021:adaattn}, IEST~\cite{chen:2021:iest} and StyleFormer~\cite{wu:2021:styleformer}.
AdaIN, Avater, SANet, AAMS, and MAST are typical CNN-based image stylization approaches.
MCC~\cite{deng:2021:arbitrary} is a video style transfer method but can be applied to images without damaging the generated results.
ArtFlow~\cite{an:2021:artflow} designs a flow-based network to minimize image reconstruction error and recovery bias.
AdaAttN~\cite{liu:2021:adaattn} performs attentive normalization on a per-point basis for feature distribution alignment.
IEST~\cite{chen:2021:iest} takes advantage of contrastive learning and external memory to boost visual quality.
StyleFormer~\cite{wu:2021:styleformer} adopts the transformer mechanism into the traditional CNN-based encoder-decoder pipeline.
By contrast, we present a \textit{pure} transformer-based architecture to solve the issue of missing content details caused by convolutions.

\paragraph{Timing information}
Our model is trained on two NVIDIA Tesla P100 GPUs and two NVIDIA GeForce RTX 3090 GPUs for approximately one day.
In Table~\ref{tab:time}, we compare the inference time of different style transfer methods under two output resolutions using one Tesla P100.

\paragraph{Qualitative evaluation}
Figure \ref{fig:compare} shows the visual results of qualitative comparisons.
Owing to the simplified alignment of mean and variance, the results of AdaIN~\cite{Huang:2017:Arbitrary} have insufficient style patterns.
The stylized images present crack artifacts that affect the overall transfer quality.
AAMS~\cite{yao:2019:attention} focuses on the main structure (referring to salient regions in the attention map) of the content image but ignores the other parts.
Therefore, the secondary structures are not well maintained.
The patch-swap-based method leads to artifacts of over-blurry output.
MCC~\cite{deng:2021:arbitrary} uses a transform formulation of self-attention, but the absence of non-linear operation limiting the maximum value of network output results in an overflow issue around object boundaries.
The flow-based model has limited capability of feature representation, thus the results of ArtFlow~\cite{an:2021:artflow} generally have the problem of insufficient or inaccurate style.
The border of stylized images may present undesirable patterns due to numerical overflow.
The per-point basis of AdaAttN~\cite{liu:2021:adaattn} leads to style degeneration, thus the stylized patterns in the generated results are not consistent with the input reference.
The visual quality of IEST~\cite{chen:2021:iest} outperforms other approaches. However, the style of generated results may not be consistent with the input style reference (the \engordnumber{1} and \engordnumber{3} rows).
Following the CNN-based ``encoder-decoder'' pipeline, results of StyleFormer~\cite{wu:2021:styleformer} still tend to missing details.
By contrast, StyTr$^2$ leverages a transformer-based network, which has better feature representation to capture long-range dependencies of input image features and to avoid missing of content and style details.
Therefore, our results can achieve well-preserved content structures and desirable style patterns.




\paragraph{Quantitative evaluation}
We calculate the content difference between the generated results and input content images as well as the style difference between the generated results and input style images, as two indirect metrics of the style transfer quality.
Intuitively, the smaller the difference the better the input content/style is preserved.
We randomly select $40$ style images and $20$ content images to generate $800$ stylized images.
For each method, we compute the content difference based on Eq.~(\ref{fun:contentloss}) and calculate the style difference following Eq.~(\ref{fun:styleloss}).
Table~\ref{tab:quancomp} shows the corresponding quantitative results.
Overall, our method achieves the lowest content losses and IEST~\cite{chen:2021:iest} is the second-best.
However, as discussed in the qualitative evaluation above, the style loss of IEST is the highest because the style appearance of generated results is not far from the input style reference.
In terms of style loss, SANet~\cite{park:2019:arbitrary} and StyTr$^2$ outperform the other methods.
Therefore, our results can effectively preserve both the input content and the reference style simultaneously.


\newcommand\biascomparisonfigurewidth{0.073}
\newcommand\biascomparisontextwidth{0.015}

\begin{figure*}
\centering
\begin{minipage}{\biascomparisontextwidth\linewidth}
    \centering
    \rotatebox{90}{\figtext{Content}}
\end{minipage}
\begin{minipage}{\biascomparisonfigurewidth\linewidth}
    \includegraphics[width=\linewidth]{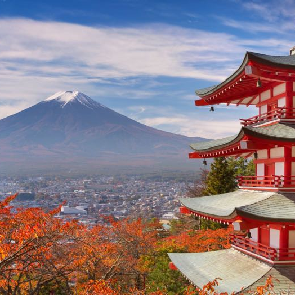}
\end{minipage}
\begin{minipage}{\biascomparisonfigurewidth\linewidth}
    \includegraphics[width=\linewidth]{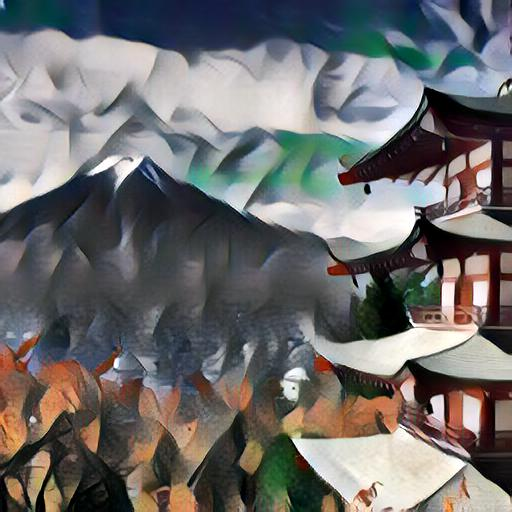}
\end{minipage}
\begin{minipage}{\biascomparisonfigurewidth\linewidth}
    \includegraphics[width=\linewidth]{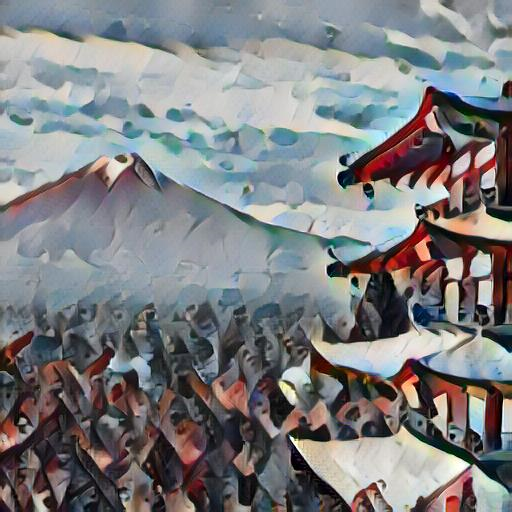}
\end{minipage}
\begin{minipage}{\biascomparisonfigurewidth\linewidth}
    \includegraphics[width=\linewidth]{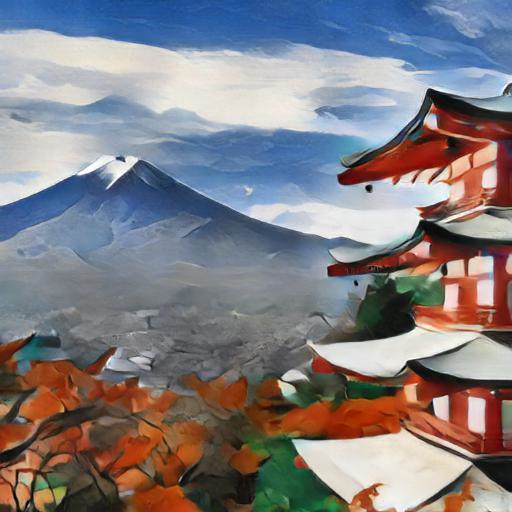}
\end{minipage}
\begin{minipage}{\biascomparisonfigurewidth\linewidth}
    \includegraphics[width=\linewidth]{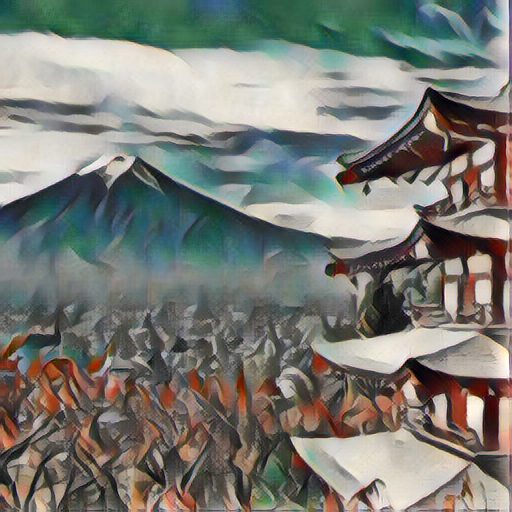}
\end{minipage}
\begin{minipage}{\biascomparisonfigurewidth\linewidth}
    \includegraphics[width=\linewidth]{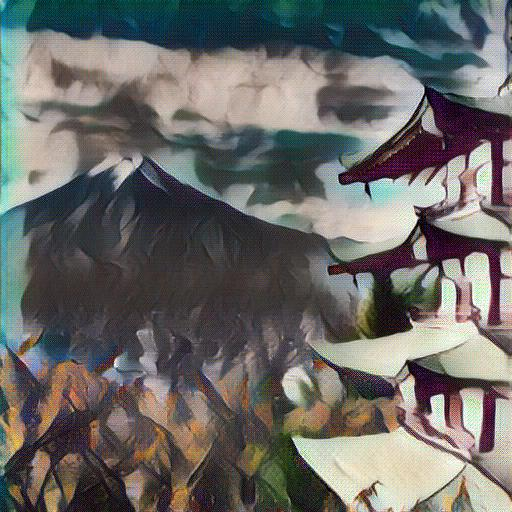}
\end{minipage}
\begin{minipage}{\biascomparisonfigurewidth\linewidth}
    \includegraphics[width=\linewidth]{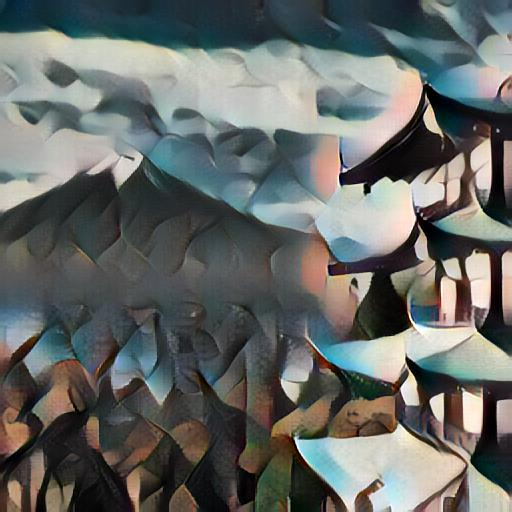}
\end{minipage}
\begin{minipage}{\biascomparisonfigurewidth\linewidth}
    \includegraphics[width=\linewidth]{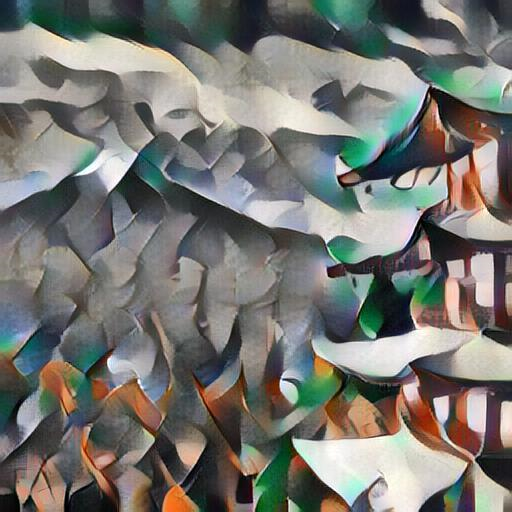}
\end{minipage}
\begin{minipage}{\biascomparisonfigurewidth\linewidth}
    \includegraphics[width=\linewidth]{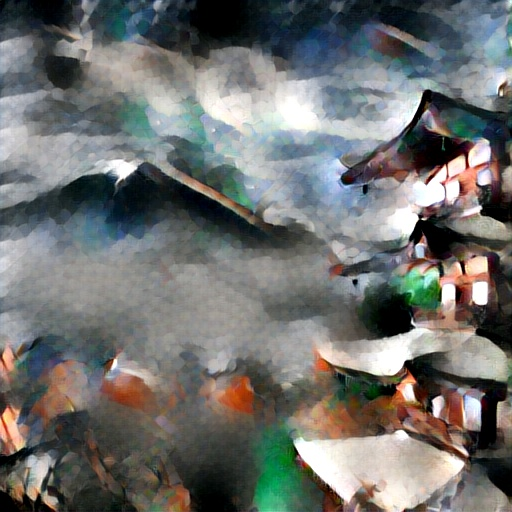}
\end{minipage}
\begin{minipage}{\biascomparisonfigurewidth\linewidth}
    \includegraphics[width=\linewidth]{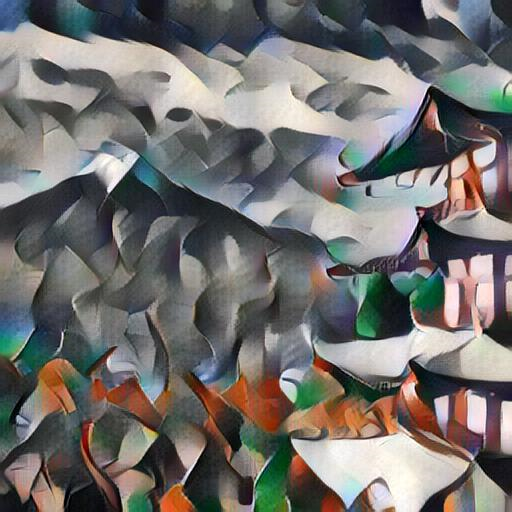}
\end{minipage}
\begin{minipage}{\biascomparisonfigurewidth\linewidth}
    \includegraphics[width=\linewidth]{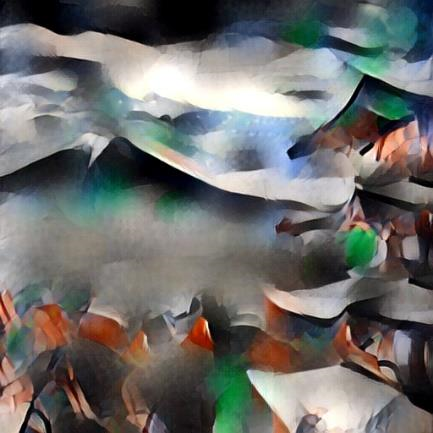}
\end{minipage}
\begin{minipage}{\biascomparisonfigurewidth\linewidth}
    \includegraphics[width=\linewidth]{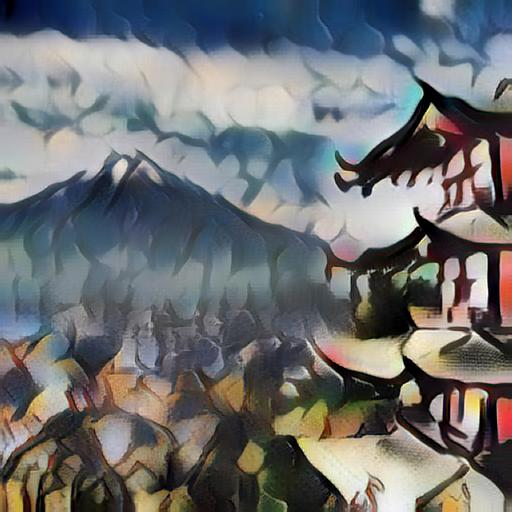}
\end{minipage}
\begin{minipage}{\biascomparisontextwidth\linewidth}
    \centering
    \rotatebox{90}{\figtext{Round 1}}
\end{minipage}

\begin{minipage}{\biascomparisontextwidth\linewidth}
    \centering
    \rotatebox{90}{\figtext{Style}}
\end{minipage}
\begin{minipage}{\biascomparisonfigurewidth\linewidth}
    \includegraphics[width=\linewidth]{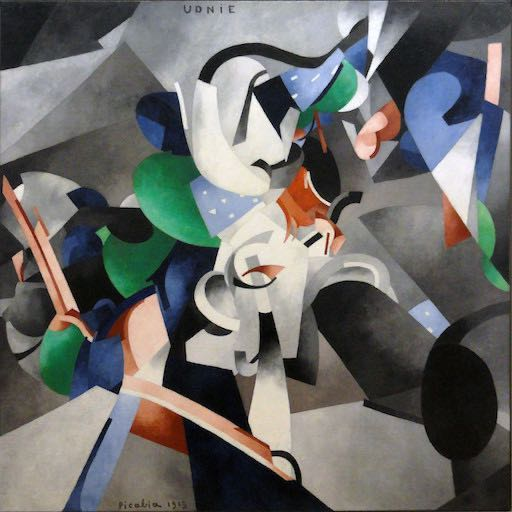}
\end{minipage}
\begin{minipage}{\biascomparisonfigurewidth\linewidth}
    \includegraphics[width=\linewidth]{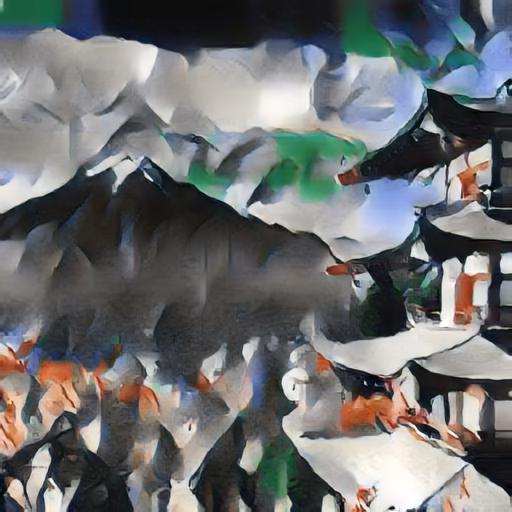}
\end{minipage}
\begin{minipage}{\biascomparisonfigurewidth\linewidth}
    \includegraphics[width=\linewidth]{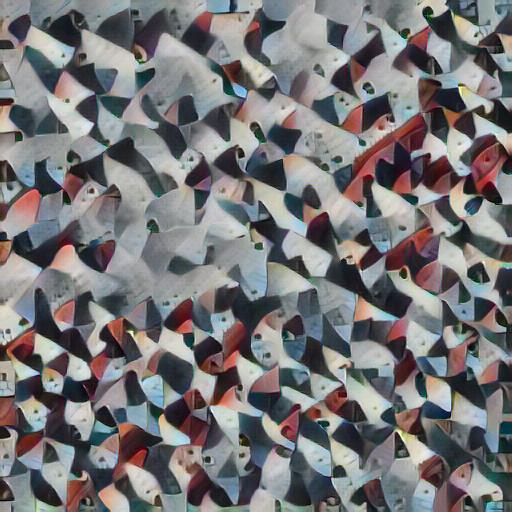}
\end{minipage}
\begin{minipage}{\biascomparisonfigurewidth\linewidth}
    \includegraphics[width=\linewidth]{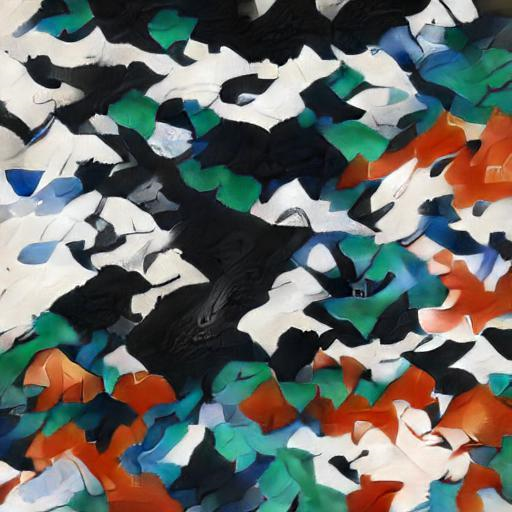}
\end{minipage}
\begin{minipage}{\biascomparisonfigurewidth\linewidth}
    \includegraphics[width=\linewidth]{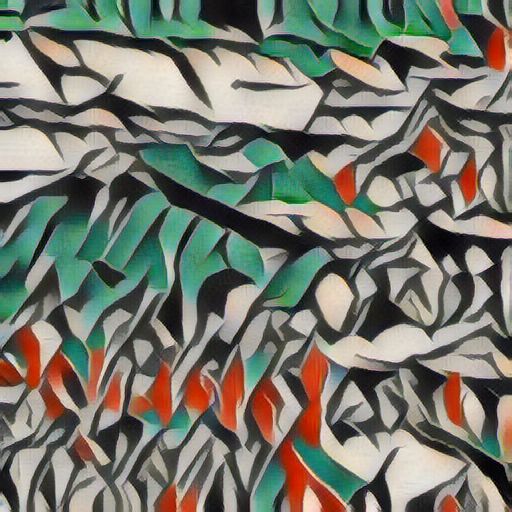}
\end{minipage}
\begin{minipage}{\biascomparisonfigurewidth\linewidth}
    \includegraphics[width=\linewidth]{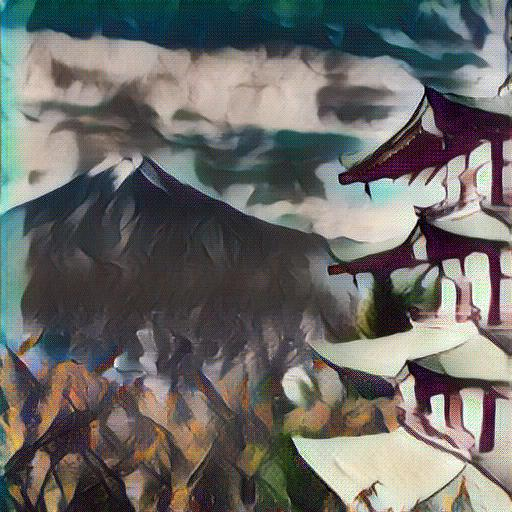}
\end{minipage}
\begin{minipage}{\biascomparisonfigurewidth\linewidth}
    \includegraphics[width=\linewidth]{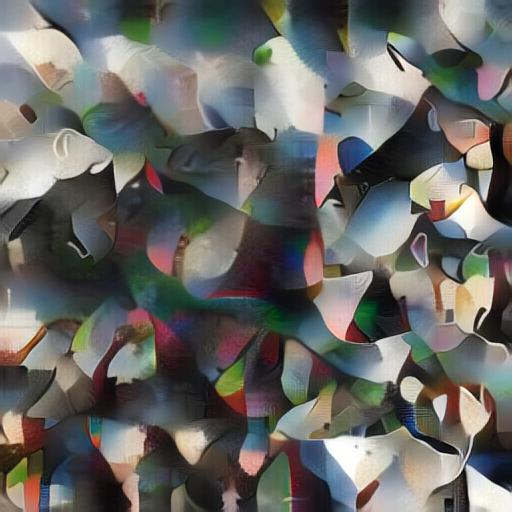}
\end{minipage}
\begin{minipage}{\biascomparisonfigurewidth\linewidth}
    \includegraphics[width=\linewidth]{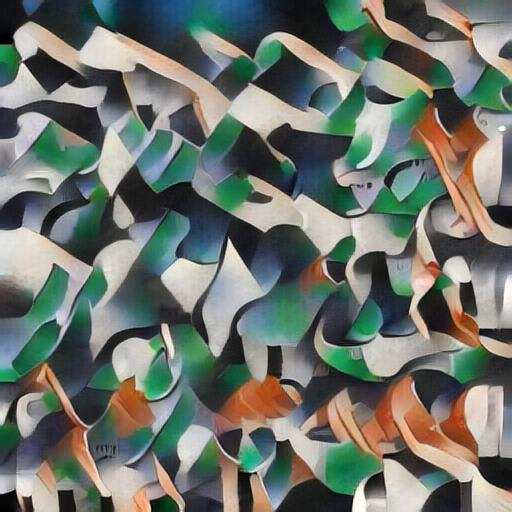}
\end{minipage}
\begin{minipage}{\biascomparisonfigurewidth\linewidth}
    \includegraphics[width=\linewidth]{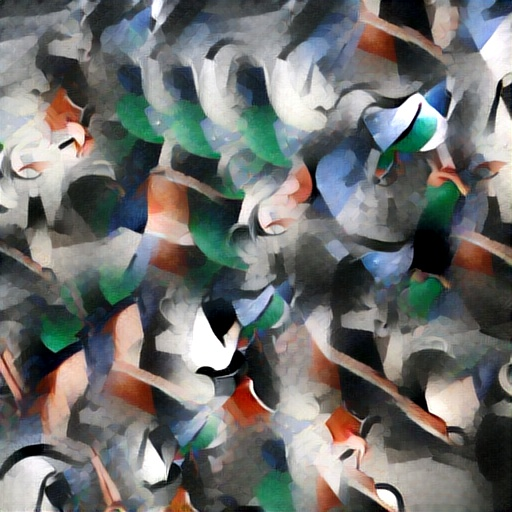}
\end{minipage}
\begin{minipage}{\biascomparisonfigurewidth\linewidth}
    \includegraphics[width=\linewidth]{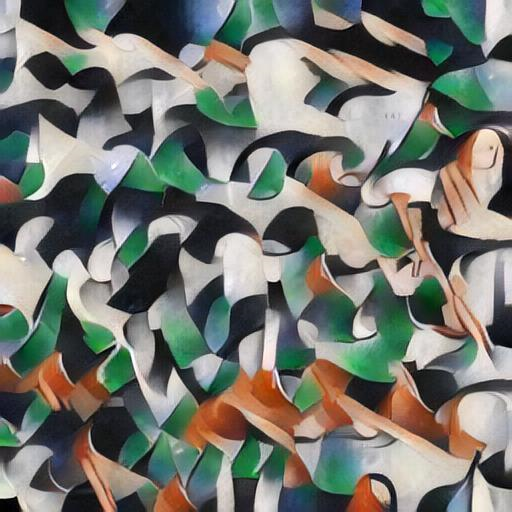}
\end{minipage}
\begin{minipage}{\biascomparisonfigurewidth\linewidth}
    \includegraphics[width=\linewidth]{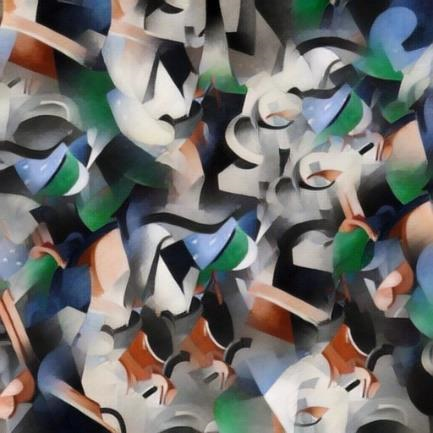}
\end{minipage}
\begin{minipage}{\biascomparisonfigurewidth\linewidth}
    \includegraphics[width=\linewidth]{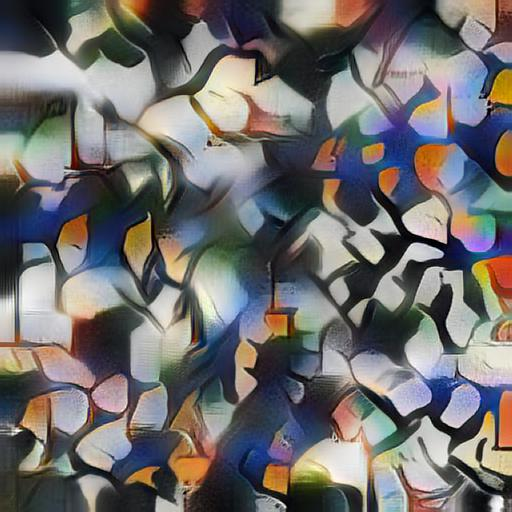}
\end{minipage}
\begin{minipage}{\biascomparisontextwidth\linewidth}
    \centering
    \rotatebox{90}{\figtext{Round 20}}
\end{minipage}


\hspace{0.07\linewidth}
\begin{minipage}{\biascomparisonfigurewidth\linewidth}
    \centering
    \figtext{Ours}
\end{minipage}
\begin{minipage}{\biascomparisonfigurewidth\linewidth}
    \centering
    \figtext{StyleFormer}
\end{minipage}
\begin{minipage}{\biascomparisonfigurewidth\linewidth}
    \centering
    \figtext{IEST}
\end{minipage}
\begin{minipage}{\biascomparisonfigurewidth\linewidth}
    \centering
    \figtext{AdaAttN}
\end{minipage}
\begin{minipage}{\biascomparisonfigurewidth\linewidth}
    \centering
    \figtext{ArtFlow}
\end{minipage}
\begin{minipage}{\biascomparisonfigurewidth\linewidth}
    \centering
    \figtext{MCC}
\end{minipage}
\begin{minipage}{\biascomparisonfigurewidth\linewidth}
    \centering
    \figtext{MAST}
\end{minipage}
\begin{minipage}{\biascomparisonfigurewidth\linewidth}
    \centering
    \figtext{AAMS}
\end{minipage}
\begin{minipage}{\biascomparisonfigurewidth\linewidth}
    \centering
    \figtext{SANet}
\end{minipage}
\begin{minipage}{\biascomparisonfigurewidth\linewidth}
    \centering
    \figtext{Avatar}
\end{minipage}
\begin{minipage}{\biascomparisonfigurewidth\linewidth}
    \centering
    \figtext{AdaIN}
\end{minipage}
\caption{Visualization of the content leak issue. Top/bottom: results after running the same stylization process using a certain method after the \engordnumber{1} and the \engordnumber{20} round, respectively.}
\label{fig:content_leak}
\end{figure*}

\subsection{Analysis of Content Leak}

The content leak issue usually occurs in the stylization process because CNN-based feature representation may not sufficiently capture details in the image content.
This type of artifact is easy to spot by human eyes after repeating several rounds of the same stylization process~\cite{an:2021:artflow}, which is formulated by
\begin{equation}
\begin{split}
\outputimage^{i} = \generator_i(\dots\generator_2(\generator_1(\inputcontent, \inputstyle), \inputstyle)\dots,\inputstyle),
\end{split}
\end{equation}
where $\generator_i$ is the generator for the $i$-th round and $\outputimage^{i}$ is the corresponding stylization result.
To solve the content leak problem, An et al~\cite{an:2021:artflow} propose a reversible network to replace CNN-based models.
However, strict reversibility may not be suitable for generation tasks~\cite{park:2020:cut}.
Furthermore, the robustness and generated visual effects of ArtFlow may be downgraded due to limited capability of feature representation.
By contrast, we leverage the capability of transformer-based architecture to capture long-range dependencies. Thus, our method can significantly alleviate the content leak issue.

We compare StyTr$^2$ with CNN-based methods and flow-based model ArtFlow~\cite{an:2021:artflow}.
Figure \ref{fig:content_leak} demonstrates the corresponding results after the 1st and the 20th rounds of the repeating stylization process.
As shown in the top row, the content structures generated by CNN-based methods after the first round are damaged to various degrees, but our result still presents clear content details.
Although the results generated by ArtFlow maintain clear content structures, the stylized effect is not satisfactory (e.g., the marginal flaws and inappropriate style patterns).
The bottom row of Figure \ref{fig:content_leak} shows that with increasing rounds of the stylization process, the content structures generated by CNN-based methods tend to be blurry, while the content structures generated by our method are remain distinct.
The same problem applies to StyleFormer, which also relies on the CNN-based encoder-decoder pipeline.
Therefore, our model captures precise content representation leading to superior style transfer results while effectively alleviating the content leak issue.

\subsection{Analysis of CAPE}
As described in \ref{sec:method_cape}, when calculating PE, we should take the semantic information of content images into account.
To compare the proposed CAPE with sinusoidal PE which is not semantics-aware, we show two cases where the input content image has repetitive patterns or is simply collaged by repeating one image four times.
As shown in Figure \ref{fig:pure}, we can observe inconsistent stylized regions in the final results when using sinusoidal PE.
The input resolution is set to be $256 \times 256$, which is the same as the image resolution for training. 

\newcommand\pecomparisonfigurewidth{0.24}

\begin{figure}
\setlength{\abovecaptionskip}{2mm}
\centering
\includegraphics[width=\pecomparisonfigurewidth\linewidth]{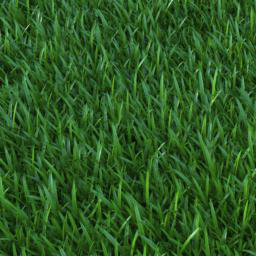}
\includegraphics[width=\pecomparisonfigurewidth\linewidth]{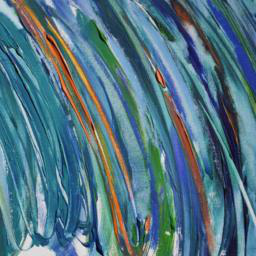}
\includegraphics[width=\pecomparisonfigurewidth\linewidth]{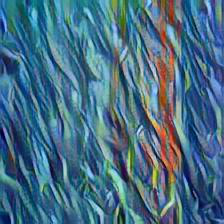}
\includegraphics[width=\pecomparisonfigurewidth\linewidth]{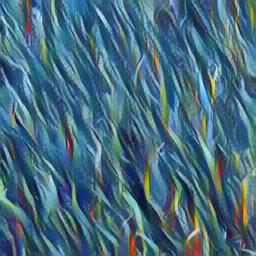}

\includegraphics[width=\pecomparisonfigurewidth\linewidth]{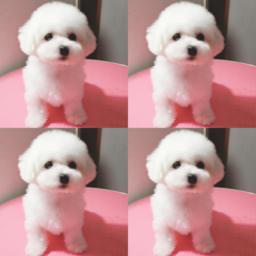}
\includegraphics[width=\pecomparisonfigurewidth\linewidth]{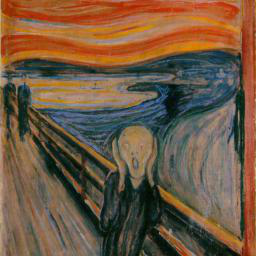}
\includegraphics[width=\pecomparisonfigurewidth\linewidth]{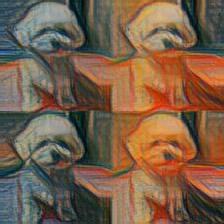}
\includegraphics[width=\pecomparisonfigurewidth\linewidth]{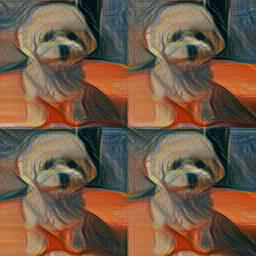}

\begin{minipage}{\pecomparisonfigurewidth\linewidth}
    \centering
    \figtext{Content}
\end{minipage}
\begin{minipage}{\pecomparisonfigurewidth\linewidth}
    \centering
    \figtext{Style}
\end{minipage}
\begin{minipage}{\pecomparisonfigurewidth\linewidth}
    \centering
    \figtext{Sinusoidal PE}
\end{minipage}
\begin{minipage}{\pecomparisonfigurewidth\linewidth}
    \centering
    \figtext{CAPE}
\end{minipage}
\caption{Comparisons of sinusoidal PE and CAPE using content images with repetitive patterns.
}
\label{fig:pure}
\end{figure}

\begin{figure*}
\setlength{\abovecaptionskip}{2mm}
\centering
\includegraphics[width=\textwidth]{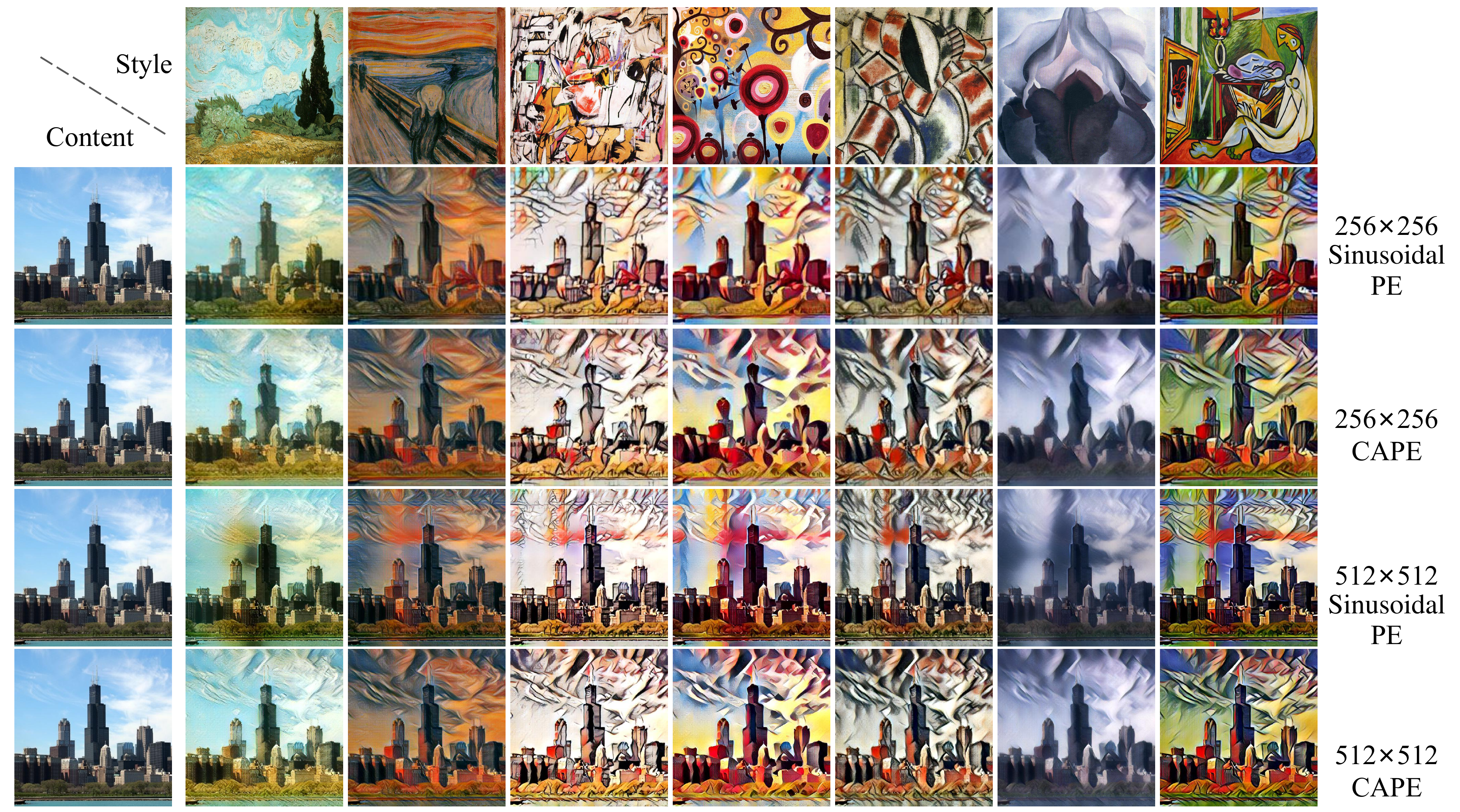}
\vspace{-15pt}
\caption{Comparisons of sinusoidal PE and CAPE using content images with two different resolutions.
}
\label{fig:position_results}
\end{figure*}

Moreover, handling input resolution different from the training examples is generally challenging for a learning-based method.
To this end, an ideal PE for vision tasks should be scale-invariant, but a drastic change of image resolution leads to a significant difference in traditional PE.
We compare our CAPE with sinusoidal PE in Figure \ref{fig:position_results}.
In the third row, the input size is $512 \times 512$, which is twice the image resolution for training.
Consequently, the results present vertical track artifacts due to the large positional deviation.
In the second row, the input resolution is $256 \times 256$, which is the same as the training data.
The corresponding results do not have the issue of vertical tracks but are not satisfactory due to the small resolution.
By contrast, our method supports any input resolution with CAPE by design.
Therefore, our results in the last row of Figure \ref{fig:position_results} present clear content structures and proper stylized patterns.
Additional ablation studies are provided in our supplementary materials.

\begin{figure}
    \setlength{\abovecaptionskip}{2mm}
    \centering
    \begin{minipage}{\linewidth}
        \includegraphics[width=0.24\linewidth]{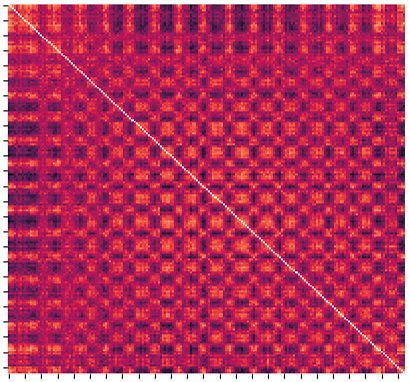}
        \includegraphics[width=0.24\linewidth]{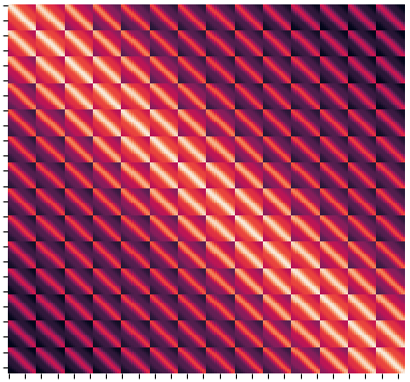}
        \includegraphics[width=0.24\linewidth]{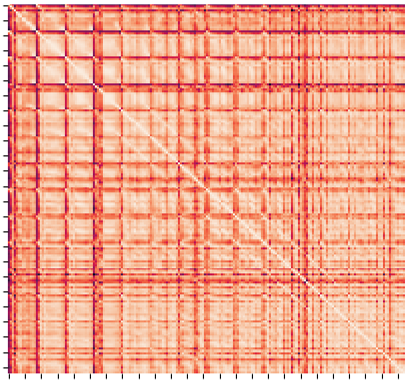}
        \includegraphics[width=0.24\linewidth]{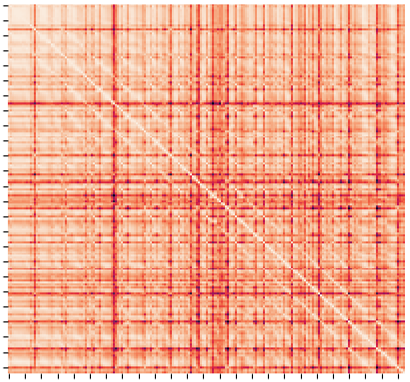}
    \end{minipage}
    \vspace{-1pt}
    \begin{minipage}{0.24\linewidth}
        \centering
        \figtext{Learnable PE}
    \end{minipage}
    \begin{minipage}{0.24\linewidth}
        \centering
        \figtext{Sinusoidal PE}
    \end{minipage}
    \begin{minipage}{0.48\linewidth}
        \centering
        \figtext{CAPE}
    \end{minipage}

    \caption{Normalized attention scores of different types of PE.}
    \label{fig:stytr2_cape_visual}
\end{figure}

To verify whether CAPE could provide position information, we show CAPEs with different inputs in Figure \ref{fig:stytr2_cape_visual}.
Although two CAPEs are not the same, they have similar encoding behaviors such as highlighted diagonal, repeat, and periodic patterns as learnable PE~\cite{dosovitskiy2021vit} and sinusoidal PE.
Different from learnable PE where the encoding is conditioned on the whole dataset, our CAPE dynamically encodes different input and thus can easily generalize to various resolutions.

\subsection{User study}
We conduct a user study to further compare our method.
AdaAttN~\cite{liu:2021:adaattn}, ArtFlow~\cite{an:2021:artflow}, MCC~\cite{deng:2021:arbitrary}, AAMS~\cite{yao:2019:attention} and  AdaIN~\cite{Huang:2017:Arbitrary} are selected as baselines.
We reuse the images in the quantitative comparison and invite $100$ participants to evaluate the results of different approaches.
The participants are informed of the purpose and details of our user study.
The participants comprise $52$ males and $48$ females, in the age range of 19$\sim$45.
Given a content image and a style image, we show the result generated by our approach and the output from another randomly selected method for comparison and ask the user to choose (1) which result has better stylization effects (2) which stylization result preserves the content structures better and (3) which stylization result transfers the style patterns more consistently.
No time limit is given for the selection process, but the assessment time should be longer than $30$ seconds for each question.
Each participant conducts $40$ rounds of comparisons and we collect $4,000$ votes for each question.
We count the votes that existing methods are preferred to ours and show the statistical results in \ref{tab:user_study}.
Our method is superior to other approaches in all three criteria of overall quality, content preservation, and style consistency.
\begin{table}
\centering
\setlength{\tabcolsep}{4.5pt}
\begin{tabular}{l|ccccc}
    \toprule
    &AdaAttN&ArtFlow & MCC  & AAMS & AdaIN \\
    \midrule
    Overall & 44.8\%&25.7\% & 25.5\%  & 18.2\% & 23.5\% \\
    \midrule
    Content & 45.7\%&28.3\% & 20.6\% & 19.4\%&  31.7\% \\
    \midrule
    Style & 23.6\%&16.7\% & 27.8\% & 14.8\% & 13.5\% \\
    \bottomrule
\end{tabular}
\vspace{-5pt}
\caption{User study results. Each number represents the percentage of votes that the corresponding method is preferred to ours, using the criteria of overall quality, preservation of content and style, respectively.}
\label{tab:user_study}
\vspace{-10pt}
\end{table}

\section{Conclusion}

In this work, we propose a novel framework called StyTr$^2$, for image style transfer.
Our StyTr$^2$ includes a content transformer encoder and a style transformer encoder to capture domain-specific long-range information.
A transformer decoder is developed to translate the content sequences based on the reference style sequences.
We also propose a content-aware positional encoding scheme that is semantics-aware and is suitable for scale-invariant visual generation tasks.
As the first baseline for style transfer using a visual transformer, StyTr$^2$ alleviates the content leak problem of CNN-based models and provides fresh insight into the challenging style transfer problem.
At present, the test-time speed of our method is not as fast as some CNN-based approaches.
Incorporating some priors from CNNs to speed up the computation would be an interesting future approach.
\vspace{-2mm}

\ifCLASSOPTIONcaptionsoff
  \newpage
\fi

\bibliographystyle{IEEEtran}
\bibliography{StyTr}

\begin{thebibliography}{10}
\providecommand{\url}[1]{#1}
\csname url@samestyle\endcsname
\providecommand{\newblock}{\relax}
\providecommand{\bibinfo}[2]{#2}
\providecommand{\BIBentrySTDinterwordspacing}{\spaceskip=0pt\relax}
\providecommand{\BIBentryALTinterwordstretchfactor}{4}
\providecommand{\BIBentryALTinterwordspacing}{\spaceskip=\fontdimen2\font plus
\BIBentryALTinterwordstretchfactor\fontdimen3\font minus
  \fontdimen4\font\relax}
\providecommand{\BIBforeignlanguage}[2]{{%
\expandafter\ifx\csname l@#1\endcsname\relax
\typeout{** WARNING: IEEEtran.bst: No hyphenation pattern has been}%
\typeout{** loaded for the language `#1'. Using the pattern for}%
\typeout{** the default language instead.}%
\else
\language=\csname l@#1\endcsname
\fi
#2}}
\providecommand{\BIBdecl}{\relax}
\BIBdecl

\bibitem{efros:2001:image}
A.~A. Efros and W.~T. Freeman, ``Image quilting for texture synthesis and
  transfer,'' in \emph{Proceedings of Annual Conference on Computer Graphics
  and Interactive Techniques}, 2001, pp. 341--346.

\bibitem{bruckner:2007:style}
S.~Bruckner and M.~E. Gr{\"o}ller, ``Style transfer functions for illustrative
  volume rendering,'' \emph{Computer Graphics Forum}, vol.~26, no.~3, pp.
  715--724, 2007.

\bibitem{gatys:2016:image}
L.~A. Gatys, A.~S. Ecker, and M.~Bethge, ``Image style transfer using
  convolutional neural networks,'' in \emph{IEEE/CVF Conference on Computer
  Vision and Pattern Recognition (CVPR)}, 2016, pp. 2414--2423.

\bibitem{li:2017:demystifying}
Y.~Li, N.~Wang, J.~Liu, and X.~Hou, ``Demystifying neural style transfer,'' in
  \emph{International Joint Conference on Artificial Intelligence (IJCAI)},
  2017.

\bibitem{risser:2017:stable}
E.~Risser, P.~Wilmot, and C.~Barnes, ``Stable and controllable neural texture
  synthesis and style transfer using histogram losses,'' \emph{arXiv preprint
  arXiv:1701.08893}, 2017.

\bibitem{Huang:2017:Arbitrary}
X.~Huang and B.~Serge, ``Arbitrary style transfer in real-time with adaptive
  instance normalization,'' in \emph{IEEE International Conference on Computer
  Vision (ICCV)}, 2017, pp. 1501--1510.

\bibitem{li:2017:universal}
Y.~Li, C.~Fang, J.~Yang, Z.~Wang, X.~Lu, and M.-H. Yang, ``Universal style
  transfer via feature transforms,'' in \emph{Advances Neural Information
  Processing Systems (NeurIPS)}, 2017, pp. 386--396.

\bibitem{wang:2020:diversified}
Z.~Wang, L.~Zhao, H.~Chen, L.~Qiu, Q.~Mo, S.~Lin, W.~Xing, and D.~Lu,
  ``Diversified arbitrary style transfer via deep feature perturbation,'' in
  \emph{IEEE/CVF Conference on Computer Vision and Pattern Recognition (CVPR)},
  2020, pp. 7789--7798.

\bibitem{li:2018:closed}
Y.~Li, M.-Y. Liu, X.~Li, M.-H. Yang, and J.~Kautz, ``A closed-form solution to
  photorealistic image stylization,'' in \emph{European Conference on Computer
  Vision (ECCV)}, 2018, pp. 453--468.

\bibitem{2021:Lin:Drafting}
T.~Lin, Z.~Ma, F.~Li, D.~He, X.~Li, E.~Ding, N.~Wang, J.~Li, and X.~Gao,
  ``Drafting and revision: Laplacian pyramid network for fast high-quality
  artistic style transfer,'' in \emph{IEEE/CVF Conference on Computer Vision
  and Pattern Recognition (CVPR)}, 2021.

\bibitem{2020:An:Real}
J.~An, T.~Li, H.~Huang, L.~Shen, X.~Wang, Y.~Tang, J.~Ma, W.~Liu, and J.~Luo,
  ``Real-time universal style transfer on high-resolution images via
  zero-channel pruning,'' \emph{arXiv preprint arXiv:2006.09029}, 2020.

\bibitem{2019:Lu:A}
M.~Lu, H.~Zhao, A.~Yao, Y.~Chen, F.~Xu, and L.~Zhang, ``A closed-form solution
  to universal style transfer,'' \emph{IEEE/CVF International Conference on
  Computer Vision (ICCV)}, 2019.

\bibitem{2019:An:Ultrafast}
J.~An, H.~Xiong, J.~Huan, and J.~Luo, ``Ultrafast photorealistic style transfer
  via neural architecture search,'' in \emph{AAAI Conference on Artificial
  Intelligence (AAAI)}, vol.~34, 2020, pp. 10\,443--10\,450.

\bibitem{2020:Wang:Collaborative}
H.~Wang, Y.~Li, Y.~Wang, H.~Hu, and M.-H. Yang, ``Collaborative distillation
  for ultra-resolution universal style transfer,'' in \emph{IEEE/CVF Conference
  on Computer Vision and Pattern Recognition}, 2020, pp. 1860--1869.

\bibitem{park:2019:arbitrary}
D.~Y. Park and K.~H. Lee, ``Arbitrary style transfer with style-attentional
  networks,'' in \emph{IEEE/CVF Conference on Computer Vision and Pattern
  Recognition (CVPR)}, 2019, pp. 5880--5888.

\bibitem{deng:2020:arbitrary}
Y.~Deng, F.~Tang, W.~Dong, W.~Sun, F.~Huang, and C.~Xu, ``Arbitrary style
  transfer via multi-adaptation network,'' in \emph{ACM International
  Conference on Multimedia}, 2020, pp. 2719--2727.

\bibitem{deng:2021:arbitrary}
Y.~Deng, F.~Tang, W.~Dong, H.~Huang, C.~Ma, and C.~Xu, ``Arbitrary video style
  transfer via multi-channel correlation,'' in \emph{AAAI Conference on
  Artificial Intelligence (AAAI)}, 2021, pp. 1210--1217.

\bibitem{yao:2019:attention}
Y.~Yao, J.~Ren, X.~Xie, W.~Liu, Y.-J. Liu, and J.~Wang, ``Attention-aware
  multi-stroke style transfer,'' in \emph{IEEE/CVF Conference on Computer
  Vision and Pattern Recognition (CVPR)}, 2019, pp. 1467--1475.

\bibitem{liu:2021:adaattn}
S.~Liu, T.~Lin, D.~He, F.~Li, M.~Wang, X.~Li, Z.~Sun, Q.~Li, and E.~Ding,
  ``Adaattn: Revisit attention mechanism in arbitrary neural style transfer,''
  in \emph{Proceedings of the IEEE International Conference on Computer
  Vision}, 2021.

\bibitem{jiang:2021:transgan}
Y.~Jiang, S.~Chang, and Z.~Wang, ``Transgan: Two transformers can make one
  strong gan,'' \emph{arXiv preprint arXiv:2102.07074}, 2021.

\bibitem{an:2021:artflow}
J.~An, S.~Huang, Y.~Song, D.~Dou, W.~Liu, and J.~Luo, ``{ArtFlow}: Unbiased
  image style transfer via reversible neural flows,'' in \emph{IEEE/CVF
  Conferences on Computer Vision and Pattern Recognition (CVPR)}, 2021, pp.
  862--871.

\bibitem{vaswani:2017:attention}
A.~Vaswani, N.~Shazeer, N.~Parmar, J.~Uszkoreit, L.~Jones, A.~N. Gomez,
  L.~Kaiser, and I.~Polosukhin, ``Attention is all you need,'' in
  \emph{Advances in Neural Information Processing Systems (NeurIPS)}, 2017.

\bibitem{paul:2021:vision}
S.~Paul and P.-Y. Chen, ``Vision transformers are robust learners,''
  \emph{arXiv preprint arXiv:2105.07581}, 2021.

\bibitem{raghu:2021:do}
M.~Raghu, T.~Unterthiner, S.~Kornblith, C.~Zhang, and A.~Dosovitskiy, ``Do
  vision transformers see like convolutional neural networks?'' \emph{arXiv
  preprint arXiv:2108.08810}, 2021.

\bibitem{johnson:2016:perceptual}
J.~Johnson, A.~Alahi, and L.~Fei-Fei, ``Perceptual losses for real-time style
  transfer and super-resolution,'' in \emph{European Conference on Computer
  Vision (ECCV)}.\hskip 1em plus 0.5em minus 0.4em\relax Springer, 2016, pp.
  694--711.

\bibitem{li:2016:precomputed}
C.~Li and M.~Wand, ``Precomputed real-time texture synthesis with markovian
  generative adversarial networks,'' in \emph{European Conference on Computer
  Vision (ECCV)}, 2016, pp. 702--716.

\bibitem{chen:2017:stylebank}
D.~Chen, L.~Yuan, J.~Liao, N.~Yu, and G.~Hua, ``Stylebank: An explicit
  representation for neural image style transfer,'' in \emph{IEEE/CVF
  Conference on Computer Vision and Pattern Recognition (CVPR)}, 2017, pp.
  1897--1906.

\bibitem{dumoulin:2016:learned}
V.~Dumoulin, J.~Shlens, and M.~Kudlur, ``A learned representation for artistic
  style,'' in \emph{International Conference on Learning Representations
  (ICLR)}, 2016.

\bibitem{Lin:2021:DAM}
M.~Lin, F.~Tang, W.~Dong, X.~Li, C.~Xu, and C.~Ma, ``Distribution aligned
  multimodal and multi-domain image stylization,'' \emph{ACM Transactions on
  Multimedia Computing, Communications, and Applications}, vol.~17, no.~3, pp.
  96:1--96:17, 2021.

\bibitem{karras:2019:stylegan}
T.~Karras, S.~Laine, and T.~Aila, ``A style-based generator architecture for
  generative adversarial networks,'' in \emph{IEEE/CVF Conference on Computer
  Vision and Pattern Recognition (CVPR)}, 2019, pp. 4401--4410.

\bibitem{huang:2018:multimodal}
X.~Huang, M.-Y. Liu, S.~Belongie, and J.~Kautz, ``Multimodal unsupervised
  image-to-image translation,'' in \emph{European Conference on Computer Vision
  (ECCV)}, 2018, pp. 172--189.

\bibitem{wu:2020:Exchangeable}
Z.~Wu, C.~Song, Y.~Zhou, M.~Gong, and H.~Huang, ``Efanet: Exchangeable feature
  alignment network for arbitrary style transfer,'' in \emph{AAAI Conference on
  Artificial Intelligence (AAAI)}, 2020, pp. 12\,305--12\,312.

\bibitem{svoboda:2020:two}
J.~Svoboda, A.~Anoosheh, C.~Osendorfer, and J.~Masci, ``Two-stage
  peer-regularized feature recombination for arbitrary image style transfer,''
  in \emph{IEEE/CVF Conference on Computer Vision and Pattern Recognition
  (CVPR)}, 2020, pp. 13\,816--13\,825.

\bibitem{wu:2021:styleformer}
X.~Wu, Z.~Hu, L.~Sheng, and D.~Xu, ``Styleformer: Real-time arbitrary style
  transfer via parametric style composition,'' in \emph{Proceedings of the
  IEEE/CVF International Conference on Computer Vision}, 2021, pp.
  14\,618--14\,627.

\bibitem{chen:2021:iest}
H.~Chen, Z.~Wang, H.~Zhang, Z.~Zuo, A.~Li, W.~Xing, D.~Lu \emph{et~al.},
  ``Artistic style transfer with internal-external learning and contrastive
  learning,'' \emph{Advances in Neural Information Processing Systems},
  vol.~34, 2021.

\bibitem{radford:2019:language}
A.~Radford, J.~Wu, R.~Child, D.~Luan, D.~Amodei, and I.~Sutskever, ``Language
  models are unsupervised multitask learners,'' \emph{OpenAI blog}, vol.~1,
  no.~8, p.~9, 2019.

\bibitem{brown:2020:language}
T.~B. Brown, B.~Mann, N.~Ryder, M.~Subbiah, J.~Kaplan, P.~Dhariwal,
  A.~Neelakantan, P.~Shyam, G.~Sastry, A.~Askell \emph{et~al.}, ``Language
  models are few-shot learners,'' in \emph{Advances in Neural Information
  Processing Systems (NeurIPS)}, 2020, pp. 1877--1901.

\bibitem{devlin:2018:bert}
J.~Devlin, M.-W. Chang, K.~Lee, and K.~Toutanova, ``Bert: Pre-training of deep
  bidirectional transformers for language understanding,'' in \emph{Conference
  of the North American Chapter of the Association for Computational
  Linguistics: Human Language Technologies, Volume 1 (Long and Short Papers)},
  2019, pp. 4171--4186.

\bibitem{liu:2019:roberta}
Y.~Liu, M.~Ott, N.~Goyal, J.~Du, M.~Joshi, D.~Chen, O.~Levy, M.~Lewis,
  L.~Zettlemoyer, and V.~Stoyanov, ``Roberta: A robustly optimized bert
  pretraining approach,'' \emph{arXiv preprint arXiv:1907.11692}, 2019.

\bibitem{radford:2018:improving}
A.~Radford, K.~Narasimhan, T.~Salimans, and I.~Sutskever, ``Improving language
  understanding by generative pre-training,'' \emph{Preprint}, 2018.

\bibitem{dai:2019:style}
N.~Dai, J.~Liang, X.~Qiu, and X.~Huang, ``Style transformer: Unpaired text
  style transfer without disentangled latent representation,'' in \emph{Annual
  Meeting of the Association for Computational Linguistics(ACL)}, 2019.

\bibitem{xu:2022:transformers}
Y.~Xu, H.~Wei, M.~Lin, Y.~Deng, K.~Sheng, M.~Zhang, F.~Tang, W.~Dong, F.~Huang,
  and C.~Xu, ``Transformers in computational visual media: A survey,''
  \emph{Computational Visual Media}, vol.~8, no.~1, pp. 33--62, 2022.

\bibitem{carion:2020:dert}
N.~Carion, F.~Massa, G.~Synnaeve, N.~Usunier, A.~Kirillov, and S.~Zagoruyko,
  ``End-to-end object detection with transformers,'' in \emph{European
  Conference on Computer Vision (ECCV)}, 2020, pp. 213--229.

\bibitem{zhu:2020:deformabledetr}
X.~Zhu, W.~Su, L.~Lu, B.~Li, X.~Wang, and J.~Dai, ``Deformable detr: Deformable
  transformers for end-to-end object detection,'' in \emph{International
  Conference on Learning Representations (ICLR)}, 2021.

\bibitem{dai:2020:up}
Z.~Dai, B.~Cai, Y.~Lin, and J.~Chen, ``Up-detr: Unsupervised pre-training for
  object detection with transformers,'' in \emph{IEEE/CVF Conference on
  Computer Vision and Pattern Recognition (CVPR)}, 2021.

\bibitem{zheng:2021:setr}
S.~Zheng, J.~Lu, H.~Zhao, X.~Zhu, Z.~Luo, Y.~Wang, Y.~Fu, J.~Feng, T.~Xiang,
  P.~H. Torr, and L.~Zhang, ``Rethinking semantic segmentation from a
  sequence-to-sequence perspective with transformers,'' in \emph{IEEE/CVF
  Conference on Computer Vision and Pattern Recognition (CVPR)}, 2021.

\bibitem{wang:2020:end}
Y.~Wang, Z.~Xu, X.~Wang, C.~Shen, B.~Cheng, H.~Shen, and H.~Xia, ``End-to-end
  video instance segmentation with transformers,'' in \emph{IEEE/CVF Conference
  on Computer Vision and Pattern Recognition (CVPR)}, 2021.

\bibitem{dosovitskiy2021vit}
A.~Dosovitskiy, L.~Beyer, A.~Kolesnikov, D.~Weissenborn, X.~Zhai,
  T.~Unterthiner, M.~Dehghani, M.~Minderer, G.~Heigold, S.~Gelly \emph{et~al.},
  ``An image is worth 16x16 words: Transformers for image recognition at
  scale,'' in \emph{International Conference on Learning Representations
  (ICLR)}, 2021.

\bibitem{wu:2020:visual}
B.~Wu, C.~Xu, X.~Dai, A.~Wan, P.~Zhang, M.~Tomizuka, K.~Keutzer, and P.~Vajda,
  ``Visual transformers: Token-based image representation and processing for
  computer vision,'' \emph{arXiv preprint arXiv:2006.03677}, 2020.

\bibitem{chen:2020:generative}
M.~Chen, A.~Radford, R.~Child, J.~Wu, H.~Jun, D.~Luan, and I.~Sutskever,
  ``Generative pretraining from pixels,'' in \emph{International Conference on
  Machine Learning (ICML)}, 2020, pp. 1691--1703.

\bibitem{liu:2021:SwinTransformer}
Z.~Liu, Y.~Lin, Y.~Cao, H.~Hu, Y.~Wei, Z.~Zhang, S.~Lin, and B.~Guo, ``Swin
  transformer: Hierarchical vision transformer using shifted windows,'' in
  \emph{IEEE/CVF Conference on Computer Vision and Pattern Recognition (CVPR)},
  2021.

\bibitem{xuL:2021:evo}
Y.~Xu, Z.~Zhang, M.~Zhang, K.~Sheng, K.~Li, W.~Dong, L.~Zhang, C.~Xu, and
  X.~Sun, ``{Evo-ViT}: Slow-fast token evolution for dynamic vision
  transformer,'' in \emph{AAAI Conference on Artificial Intelligence (AAAI)},
  2022.

\bibitem{chen:2020:pre}
H.~Chen, Y.~Wang, T.~Guo, C.~Xu, Y.~Deng, Z.~Liu, S.~Ma, C.~Xu, C.~Xu, and
  W.~Gao, ``Pre-trained image processing transformer,'' in \emph{IEEE/CVF
  Conference on Computer Vision and Pattern Recognition (CVPR)}, 2021.

\bibitem{shaw:2018:self}
P.~Shaw, J.~Uszkoreit, and A.~Vaswani, ``Self-attention with relative position
  representations,'' in \emph{Annual Conference of the North American Chapter
  of the Association for Computational Linguistics (NAACL)}, 2018.

\bibitem{yang:2019:xlnet}
Z.~Yang, Z.~Dai, Y.~Yang, J.~Carbonell, R.~Salakhutdinov, and Q.~V. Le,
  ``Xlnet: Generalized autoregressive pretraining for language understanding,''
  in \emph{Advances in Neural Information Processing Systems (NeurIPS)}, 2019.

\bibitem{raffel:2019:exploring}
C.~Raffel, N.~Shazeer, A.~Roberts, K.~Lee, S.~Narang, M.~Matena, Y.~Zhou,
  W.~Li, and P.~J. Liu, ``Exploring the limits of transfer learning with a
  unified text-to-text transformer,'' \emph{Journal of Machine Learning
  Research}, vol.~21, no. 140, pp. 1--67, 2020.

\bibitem{he:2020:deberta}
P.~He, X.~Liu, J.~Gao, and W.~Chen, ``Deberta: Decoding-enhanced bert with
  disentangled attention,'' in \emph{International Conference on Learning
  Representations (ICLR)}, 2021.

\bibitem{xu:2020:positional}
R.~Xu, X.~Wang, K.~Chen, B.~Zhou, and C.~C. Loy, ``Positional encoding as
  spatial inductive bias in gans,'' in \emph{IEEE/CVF Conference on Computer
  Vision and Pattern Recognition (CVPR)}, 2021.

\bibitem{2020:Islam:How}
M.~A. Islam, S.~Jia, and N.~D.~B. Bruce, ``How much position information do
  convolutional neural networks encode?'' in \emph{International Conference on
  Learning Representations (ICLR)}, 2020.

\bibitem{lin:2014:coco}
T.-Y. Lin, M.~Maire, S.~Belongie, J.~Hays, P.~Perona, D.~Ramanan,
  P.~Doll{\'a}r, and C.~L. Zitnick, ``Microsoft {COCO}: Common objects in
  context,'' in \emph{European Conference on Computer Vision (ECCV)}, 2014, pp.
  740--755.

\bibitem{phillips:2011:wiki}
F.~Phillips and B.~Mackintosh, ``Wiki art gallery, inc.: A case for critical
  thinking,'' \emph{Issues in Accounting Education}, vol.~26, no.~3, pp.
  593--608, 2011.

\bibitem{kingma:2014:adam}
D.~P. Kingma and J.~Ba, ``Adam: A method for stochastic optimization,''
  \emph{arXiv preprint arXiv:1412.6980}, 2014.

\bibitem{xiong:2020:layer}
R.~Xiong, Y.~Yang, D.~He, K.~Zheng, S.~Zheng, C.~Xing, H.~Zhang, Y.~Lan,
  L.~Wang, and T.~Liu, ``On layer normalization in the transformer
  architecture,'' in \emph{International Conference on Machine Learning
  (ICML)}, 2020, pp. 10\,524--10\,533.

\bibitem{sheng2018avatar}
L.~Sheng, Z.~Lin, J.~Shao, and X.~Wang, ``Avatar-net: Multi-scale zero-shot
  style transfer by feature decoration,'' in \emph{Computer Vision and Pattern
  Recognition (CVPR), 2018 IEEE Conference on}, 2018, pp. 1--9.

\bibitem{park:2020:cut}
T.~Park, A.~A. Efros, R.~Zhang, and J.-Y. Zhu, ``Contrastive learning for
  unpaired image-to-image translation,'' in \emph{European Conference on
  Computer Vision}, 2020.

\end{thebibliography}
\newpage

\end{document}